\documentclass{article}
\usepackage[utf8]{inputenc}
\usepackage[T1]{fontenc}
\usepackage[english]{babel}
\usepackage{arxiv}
\usepackage{amsmath,amssymb}
\usepackage{graphicx}
\usepackage{booktabs,longtable,array}
\usepackage{placeins}
\usepackage{algorithm,algpseudocode}
\usepackage{microtype}
\usepackage{url}
\usepackage[authoryear,round]{natbib}
\usepackage[unicode,hidelinks,bookmarksnumbered]{hyperref}
\usepackage{cleveref}

\numberwithin{table}{section}
\numberwithin{figure}{section}
\newlength{\personatablewidth}

\title{Emergi-PersonaOS: A Persona Agent Operating System for
Situational Adaptation and Controllable Evolution}
\author{%
  Haoluan Fu\textsuperscript{1},
  Keni Chen\textsuperscript{2},
  Xinyu Jia\textsuperscript{1},
  Jinpeng Wang\textsuperscript{1}\thanks{Corresponding authors.\\
    Jinpeng Wang: \texttt{wangjinpeng@emergi.cn}\\
    Yuyu Yin: \texttt{yinyuyu@hdu.edu.cn}},
  Yuyu Yin\textsuperscript{2}\footnotemark[1],
  Yubiao Hu \textsuperscript{3} \\
  {\mdseries\textsuperscript{1}Emergi Lab of Qianzhen Digital Tech, China}\\
  {\mdseries\textsuperscript{2}Hangzhou Dianzi University, China} \\
  {\mdseries\textsuperscript{3}Electronic Soul, China}
}
\date{}
\renewcommand{\shorttitle}{Emergi-PersonaOS}
\hypersetup{%
  pdftitle={Emergi-PersonaOS: A Persona Agent Operating System for Situational Adaptation and Controllable Evolution},
  pdfauthor={Haoluan Fu, Keni Chen, Xinyu Jia, Jinpeng Wang, Yuyu Yin},
  pdflang={en}%
}

\newif\ifshowcontents
\showcontentsfalse

\begin{document}
\maketitle
\begin{abstract}
Symbiosis between humans and digital beings offers a vision for the future of human--machine interaction. In enduring human--machine relationships, personality provides a foundation for continuity of identity, individuality in interaction, and development through experience. We investigate this capacity through persona agents as computational implementations and introduce Emergi-PersonaOS, a psychology-grounded operating system for managing persona objects throughout their lifecycle. The system organizes dispositional traits, characteristic adaptations, and narrative identity into a three-layer persona representation, distinguishing relatively enduring persona beliefs from their activation in the current persona state. During situational adaptation, it integrates the current interlocutor, relationship, event, and retrieved memories to infer a persona state and generate actions and replies; during long-term development, it records experiences and outcomes, and develops and evaluates revision candidates through change attribution, meaning-making, and behavioral testing. Belief updates are managed through explicit review, traceable evidence and version records, and the ability to reject candidates, making persona evolution controllable. Using television-character dialogue as longitudinal material, we demonstrate long-horizon system operation and examine its principal mechanisms in a concrete implementation. This work provides a computational framework for persona agents to maintain individual continuity, produce situation-specific expression, and develop through experience over sustained interaction.

\mbox{}
\end{abstract}

\ifshowcontents
\tableofcontents
\clearpage
\fi

\section{Introduction}
\label{sec:introduction}

Advances in autonomy, personalization, and social interaction have enabled artificial agents to establish and maintain socio-emotional relationships with humans \citep{bickmore2005} and to provide personalized experiences in emotional companionship, educational tutoring, and customer service \citep{chen2024a}. Human--machine relationships are extending from short-term, task-specific exchanges toward long-term coexistence and sustained collaboration \citep{chen2024b}. Research has consequently begun to describe artificial entities with social capacities and a distinct individual presence as digital beings \citep{yao2025}. This development gives renewed relevance to the prospect of human--machine symbiosis \citep{licklider1960}.

The prospect of human--machine symbiosis calls for digital beings with capacities for memory, perception, and action \citep{yao2025}. Personality provides an organizing basis for maintaining identity across time, forming individualized relationships with different interaction partners, and developing through accumulated experience. We focus on this personality-related capacity, using persona agents as its technical realization. Research on persona-based role-playing agents has already incorporated identity information, character knowledge, emotions, and behavioral patterns into models, enabling expression with persistent individual characteristics \citep{shao2023}. However, a unified mechanism across time scales is still needed to explain how current interlocutors, relationships, and events give rise to a present persona state, and how successive experiences justify longer-term persona change. This need has two closely related aspects.

The first is situational adaptation. Personality research indicates that behavior reflects both relatively enduring personal characteristics and the specific situation. \citet{mischel1995} describe recurrent situation--behavior patterns organized around the psychological features of situations. More recently, \citet{harry2026} reported that situation-related within-person dynamics accounted for 72\%--74\% of variation in personalized expression in their study. Repeated access to a fixed profile therefore offers only a partial account of how the same individual responds to different interlocutors, relationships, and events. Persona agents require an explicit current state that connects enduring persona content to immediate behavior. Situational evidence, existing persona beliefs, and relevant experiences inform which content is activated, and this state then guides actions and replies. Such a mechanism can represent both continuity across situations and differences in situated expression.

The second is long-term persona updating. Research on personality development associates major life events with long-term personality change \citep{buhler2024}, and recurring situations, states, behaviors, and feedback contribute to this developmental process \citep{wrzus2017}. Existing agents preserve experiences in external memory \citep{packer2023} and use reflection to translate feedback into strategies for subsequent action \citep{shinn2023}. These mechanisms provide a technical basis for retaining information and using experience over time. An account of personality development must also specify an event's developmental significance, the individual's interpretation of it, the relationships and situations to which a change applies, and the evidence that warrants revising enduring persona content. Recent work has identified compressed individual differences and insufficiently supported directions of change in reflection-driven persona evolution \citep{wang2026b}. Persona agents designed for long-term symbiosis therefore need to identify developmentally meaningful events within accumulated experience, develop and evaluate revision candidates through change attribution, meaning-making, and behavioral testing, and manage long-term updates using event evidence and subsequent behavior.

To address these two aspects, we introduce Emergi-PersonaOS, a psychology-grounded persona agent operating system that manages persona objects throughout their lifecycle. A shared persona object connects persona construction, situational expression, and long-term persona updating. It organizes dispositional traits, characteristic adaptations, and narrative identity into a three-layer representation: persona beliefs preserve relatively enduring content, and persona states record its activation and expression in the current situation. For situational adaptation, the system holds the active belief version fixed, collects situational evidence, infers the current state across the three layers, and generates actions and replies from that state. Differences in expression are thus localized to a state representation whose basis can be inspected. For long-term persona updating, the system records events, states, behavior, and outcomes, then develops and evaluates revision candidates through change attribution, meaning-making, and behavioral testing. A candidate with sufficient event evidence and an explicitly delimited scope takes effect as a new belief version after review. Rejected and deferred candidates remain eligible for reassessment, and every revision retains source and version records. We define controllable evolution through these three properties: reviewable revision decisions, traceable evidence and versions, and rejectable candidates. Together, the two processes connect enduring content, current expression, and development over time through the same persona object and its version history. We demonstrate the complete system using longitudinal dialogue material from a television character.

This paper makes three contributions. First, we introduce a persona agent operating system for long-term human--machine symbiosis that manages the full persona lifecycle. Its three-layer representation organizes dispositional traits, characteristic adaptations, and narrative identity, while separating enduring beliefs from current states so that construction, situational expression, and long-term updating use a common representation. Second, we develop a state-based mechanism for situational adaptation. By incorporating interlocutors, relationships, events, and relevant experience, the mechanism connects enduring beliefs, current states, and concrete behavior, providing a psychologically informed and interpretable account of individualized expression. Third, we develop an event-evidence-driven mechanism for controllable persona evolution. It translates experience, change attribution, meaning-making, and behavioral testing into a long-term updating procedure with reviewable decisions, traceable evidence and versions, and rejectable candidates, supporting development alongside individual continuity.

\section{Related Work}
\label{sec:related-work}

\subsection{Personality Structure and Computational Representation}
\label{sec:source-12}

Persona-based role-playing agents commonly organize character information through profiles, experiential material, and behavioral examples. RoleLLM uses character profiles, knowledge instructions, and linguistic-style examples for character-specific training \citep{wang2024a}. Character-LLM reconstructs experiences to model character knowledge, emotions, and behavior \citep{shao2023}. CoSER expands these materials to include experiences, interaction settings, and inner thoughts \citep{wang2025}. CoPE relates specific situations, short-term states, and long-term traits in an interpretable approach to personality recognition \citep{sun2024}. These methods broaden the information available for character modeling. However, the organization and functional relationships of that information generally follow task-specific objectives. A common psychological account of its levels and roles remains important for integrated representation design.

\subsection{Situational Adaptation of Personas}
\label{sec:source-14}

Research on situational adaptation examines how persona content is selected and dynamically expressed under different conditions. PersonaForge selects character information through hierarchical profiles and dual-process generation \citep{tong2026}. ThinkPersona uses persona graphs and query analysis to organize the evidence underlying a response \citep{cai2026}. State-machine approaches regulate persona expression through scoring and state transitions \citep{pielage2026}. JPAF organizes core expression, temporary adaptation, and long-term change into a structured process \citep{wang2026a}. PHASE-Tree manages character fields and their update conditions in a multi-time-scale state tree \citep{tang2026}. With the same persona specification, different conversational contexts can systematically alter linguistic, behavioral, and emotional expression. Whole Trait Theory interprets these differences as situation-sensitive expression compatible with personality consistency \citep{han2026}. These approaches incorporate situational information into persona-conditioned generation. A more complete process account is still needed to explain how that information gives rise to a current state and its behavioral expression.

Situational judgment also depends on experiences relevant to the current task. MemGPT manages long-term interaction material through in-context and external storage \citep{packer2023}. Task-conditioned retrieval additionally considers the applicability of memories to the present task \citep{zhang2026b}. PGMem links persona information to event sources through a persona--memory graph \citep{choi2026}. DREAM draws on the activating-event--belief--consequence model to organize character experiences into an event-aware memory graph with temporal and causal links, from which it constructs a two-level profile of enduring traits and event-driven behavior \citep{xiao2026}. Situational adaptation consequently requires the retrieval of relevant experience together with provenance links between that experience and the resulting persona judgments.

Once situational information informs behavior, the correspondence between the inferred state and the expressed behavior also requires evaluation. InCharacter assesses personality fidelity through psychological interviews \citep{wang2024b}, and CharacterEval evaluates role-playing dialogue across multiple dimensions \citep{tu2024}. The reliability and validity of LLM personality measurements depend on the model and administration conditions \citep{serapio2025}; self-reported traits can also diverge from behavioral outputs \citep{han2025}. Item order, paraphrasing, reasoning mode, and conversation history further affect personality measurements in LLMs \citep{tosato2025}. These findings motivate joint recording of current states, actual behavior, and evaluation conditions when assessing how persona content contributes to expression.

\subsection{Controllable Persona Evolution}
\label{sec:source-18}

Research on long-term interaction commonly uses memory and reflection to make prior experience available for subsequent action. Generative Agents connects observational memory, reflection, and planning \citep{park2023}. Reflexion translates task feedback into verbal reflections for later action \citep{shinn2023}. TRUSTMEM checks coverage, preservation, and the faithfulness of newly added content during memory transitions \citep{yang2026}. Research on continual memory consolidation shows that repeated rewriting can reduce the utility of the original experience \citep{zhang2026a}. TOKI manages contradictions and revisions in persistent memory using event time, record time, and audit histories \citep{wang2026c}. These methods support experience preservation, feedback utilization, and historical traceability, with updates primarily targeting memory content and action strategies. Persona evolution additionally requires an account of which events justify long-term change and how that change should revise character information.

Recent persona-agent research has begun to incorporate long-term change into explicit update procedures. Verification-Gated Persona State Transitions separates rapidly updated event memory from more slowly changing persona states and uses verification to govern update commits \citep{li2026}. JPAF incorporates reflection-driven long-term persona updates into personality control \citep{wang2026a}, while PHASE-Tree assigns evidence and timing requirements to individual character fields \citep{tang2026}. BFI-Adapt identifies directional biases and compressed individual differences in model-generated personality changes after life events \citep{wang2026b}. PTCBENCH measures Big Five traits under 12 external-context and life-event conditions. Across 39,240 records, it reports significant personality shifts under conditions such as unemployment, with these shifts further altering model reasoning performance \citep{yu2026}. ArcANE evaluates behavior at different points in a character's development \citep{song2026}, and BeliefShift tracks belief changes, contradictions, and evidential support across sessions \citep{myakala2026}. CORE separates current-turn local evidence from revisions to persistent persona states and applies uncertainty-aware belief revision to selectively update user profiles \citep{zhang2026c}. PersonaTree provides evidence-to-claim support paths in a three-layer persona tree, using conservative writes and confidence-guided consolidation to maintain the agent's understanding of the user \citep{hou2026}. Evolving Agents uses behavioral reflection to develop new personality traits in a trait--behavior cycle driven by multiple days of interaction in a simulated environment; revision eligibility and activation conditions remain unspecified \citep{li2024}. These contributions advance state updating, developmental evaluation, and evidence management. Controllable evolution further calls for a continuous process connecting change attribution, revision eligibility, behavioral testing, and version activation.

System-level research provides additional architectural references. AIOS organizes scheduling, context management, storage, and access control as services for agent execution \citep{mei2024}. Mumon uses three-layer persona information for generation, memory association, and multimodal expression, organizing these capabilities as a persona agent operating system \citep{mumon2026}. Existing research addresses persona representation, situational adaptation, and long-term updating. Lifecycle-level integration additionally requires these processes to share a persona object and retain event evidence and activation records for long-term changes.

Together, these studies establish three related lines of inquiry. Character modeling represents persona content that persists across situations; situational modeling describes responses within current relationships and events; memory and updating research examines how new experiences support the preservation, revision, and replacement of existing information. The three lines address different time scales and information objects, often through separate representations and execution mechanisms. Their integration calls for an explicit account of how enduring persona content, current states, and event evidence relate, and how those relationships are recorded continuously across lifecycle stages.

\section{Overall Design of Emergi-PersonaOS}
\label{sec:system-design}

Emergi-PersonaOS supports persona construction, situational expression, and long-term development in sustained human--machine interaction through six functional modules organized around a shared persona object. Separating relatively enduring persona beliefs from the current persona state allows behavior to respond to situational changes and makes long-term revisions traceable to experience under explicit activation conditions.

\subsection{Theoretical Foundations and Design Principles}
\label{sec:foundations}

\subsubsection{Theoretical Foundations}
\label{sec:theoretical-foundations}

\paragraph{Defining and organizing persona information.}

The integrative personality framework organizes individual differences into dispositional traits, characteristic adaptations, and narrative identity \citep{mcadams2006}. Traits describe general tendencies across situations; characteristic adaptations include motivational content such as needs, values, goals, and strategies; narrative identity organizes self-understanding through life stories. The framework spans disposition, motivation, and identity, with an explicit psychological foundation for each level. Dispositional traits have established classification and measurement systems; characteristic adaptations and narrative identity can be operationalized in structured form through their corresponding theoretical constructs. Emergi-PersonaOS uses these levels as the common content structure of persona beliefs and states, defining the boundaries of each layer and providing a psychological basis for field definitions, semantic organization, and implementation.

\begin{table}[htbp]
\centering
\caption{Theoretical foundations of fields in the three-layer representation.}
\label{tab:personality-fields}
\small
\renewcommand{\arraystretch}{1.15}
\setlength{\personatablewidth}{\dimexpr\linewidth-4\tabcolsep\relax}
\begin{tabular}{@{}>{\raggedright\arraybackslash}p{0.20\personatablewidth}>{\raggedright\arraybackslash}p{0.40\personatablewidth}>{\raggedright\arraybackslash}p{0.40\personatablewidth}@{}}
\toprule
\textbf{Layer} & \textbf{Fields} & \textbf{Theoretical foundation} \\
\midrule
L3: Narrative identity & Overall identity, life course, imago prototypes & Life-story theory \\[4pt]
L2: Characteristic adaptations & Needs, values, goals, strategies & Self-determination theory, basic human values theory, personal goals theory, and coping theory \\[4pt]
L1: Dispositional traits & Big Five traits & Big Five theory \\[4pt]
\bottomrule
\end{tabular}
\end{table}

Whole Trait Theory conceptualizes traits as distributions of personality states and uses social-cognitive processes to explain state variation \citep{fleeson2015}. State-density research documents substantial within-person variation across situations \citep{fleeson2001}. The persona object draws on this distinction: beliefs preserve relatively enduring content across turns, states record its current activation and expression, and long-term revisions target the beliefs. Situational fluctuations remain in the state representation, while persistent changes undergo review before entering the beliefs. This gives enduring content a single, traceable update pathway.

Persona construction requires traceable inferences from character material. The Realistic Accuracy Model identifies relevance, availability, detection, and utilization of cues as conditions for accurate personality judgment \citep{funder1995}. Source-monitoring research describes how contextual and processing records inform judgments about the origin of information \citep{johnson1993}. The construction module records the source and degree of inference for each persona entry, distinguishing manually configured content, source-based inference, generated settings, and measurement results as separate evidence types. Subsequent review can therefore distinguish directly supported facts from system-generated inferences and trace errors to specific sources.

\paragraph{Theoretical foundations of situational adaptation.}

The cognitive-affective personality system describes recurrent situation--behavior signatures organized around the psychological features of situations \citep{mischel1995}. In the execution engine, three-layer state inference links these features to persona entries and represents the current expression of persona content. Expressions across situations can consequently be compared within the same persona. Each state component retains links to the relevant persona entries and situational evidence, making the basis of the update available for inspection.

Dual-process theory distinguishes rapid autonomous processing from analytic processing that depends on working memory \citep{evans2013}. The execution engine uses this distinction to organize generation into an initial response and an analytic review. The initial stage forms a response; the review stage checks its correspondence with the three-layer state before release. This division of functions makes discrepancies between state and expression subject to an explicit pre-output check.

For memory retrieval, the encoding-specificity principle relates retrieval effectiveness to the match between current cues and the original encoding conditions \citep{tulving1973}. Event encoding preserves the original context and retrieval cues, and the memory module selects experiences by their match to the current situation. The retrieval record identifies which experiences were selected and why, making the evidence used in situational judgment available for review.

For the associative organization of persona memory, the self-memory system organizes autobiographical knowledge into lifetime periods, general events, and event-specific knowledge, and describes the influence of the working self on retrieval \citep{conway2000}. Persona memory follows this organization when linking experience to persona entries: narrative identity links to self-defining events, characteristic adaptations to goal trajectories, and dispositional traits to behavioral evidence across situations. Each entry can consequently be traced to supporting experience. Reflection and evolution retrieve evidence by layer, and the same experience can support judgments at several layers.

For persona state monitoring, ecological momentary assessment examines within-person variation through records of situations and states close to the time of occurrence \citep{shiffman2008}. The system stores state snapshots together with actual behavior and observable outcomes, producing temporally ordered within-person trajectories. Time-indexed records allow a change to be interpreted against the surrounding behavior and outcomes, distinguishing reasonable fluctuation from persistent deviation. The records needed for within-person comparison accumulate during system operation.

\paragraph{Theoretical foundations of controllable evolution.}

Controllable evolution is implemented through persona reflection and persona evolution. Its theoretical foundations concern the organization of the developmental process and the conditions governing revision review.

For the organization of persona evolution, the TESSERA framework describes repeated cycles of triggering situations, expectancies, states or behaviors, and reactions, with associative, reinforcing, and reflective processes contributing to long-term development \citep{wrzus2017}. Event records, reflection, and belief revision are organized around this cycle, connecting short-term state and behavioral records to experience interpretation and longer-term revision. This structure separates development into inspectable stages. Revision candidates arise from recurring patterns or sufficiently contextualized major events; individual state changes remain subject to this evidential process.

For evidence review, a meta-analysis of life events reports small overall associations with personality change that vary by event domain and temporal course \citep{buhler2024}. This finding informs the requirements for long-term revision: supporting material is examined by event type, duration, and scope of applicability, and a major event's developmental implications require explicit assessment. Subjective experiences of similar events also show substantial individual differences in their associations with personality trajectories \citep{schwaba2023}. Review therefore considers the individual's existing trajectory alongside event evidence, allowing the same event to support different revision decisions for different personas.

For change attribution in reflection, the covariation model compares causal explanations using consistency over time, distinctiveness across situations, and consensus across individuals \citep{kelley1973}. Attributional theory of achievement motivation analyzes causes by locus, persistence, and controllability \citep{weiner1985}. The system combines these criteria: covariation information is first used to compare explanations involving the situation, relationship, strategy applicability, and long-term change; the selected explanation is then classified by locus, persistence, and controllability. These dimensions provide explicit criteria for comparison, and the resulting causal classification accompanies each revision candidate for evidence review and scheduling.

For meaning-making in reflection, narrative-identity research describes how difficult experiences can enter self-understanding through exploration and narrative interpretation. Revisions at the narrative-identity layer distinguish event facts, meanings attributed by the character, and system inferences. Event facts describe the experience itself, attributed meanings record the character's interpretation, and system inferences are separately labeled. Each proposed identity revision can thus be traced to the type of information from which it was derived.

These theoretical foundations inform both the data structures and the execution mechanisms. The organization of persona information defines the three-layer representation, and the distinction between enduring tendencies and short-term states defines the roles of beliefs and states. Theories of personality judgment inform persona construction; theories of situational adaptation inform the execution engine, memory, and state monitoring; and theories of development inform reflection and the review procedure for evolution. Table~\ref{tab:psychological-foundations} summarizes these correspondences.

\FloatBarrier
\begingroup
\small
\renewcommand{\arraystretch}{1.12}
\setlength{\personatablewidth}{\dimexpr\linewidth-4\tabcolsep\relax}
\setlength{\LTleft}{0pt}
\setlength{\LTright}{0pt}
\setlength{\LTcapwidth}{\linewidth}
\begin{longtable}{@{}>{\raggedright\arraybackslash}p{0.19\personatablewidth}>{\raggedright\arraybackslash}p{0.34\personatablewidth}>{\raggedright\arraybackslash}p{0.47\personatablewidth}@{}}
\caption{Correspondence between system components and psychological foundations.}\label{tab:psychological-foundations}\\
\toprule
\textbf{System function} & \textbf{Psychological foundation} & \textbf{Design contribution} \\
\midrule
\endfirsthead
\multicolumn{3}{l}{\tablename\ \thetable{} (continued)}\\
\toprule
\textbf{System function} & \textbf{Psychological foundation} & \textbf{Design contribution} \\
\midrule
\endhead
\midrule
\multicolumn{3}{r}{Continued on next page}\\
\endfoot
\bottomrule
\endlastfoot
Persona structure & Three-layer personality structure \citep{mcadams2006}; Whole Trait Theory \citep{fleeson2015} & A common three-layer structure represents enduring content and its expression at the current turn. \\[4pt]
Persona construction & Realistic Accuracy Model \citep{funder1995}; source monitoring \citep{johnson1993} & Sources and degrees of inference remain explicit from source material to persona judgment. \\[4pt]
Situational interpretation & Cognitive-affective personality system \citep{mischel1995} & Situational links connect enduring content to current states and document the basis of expression differences. \\[4pt]
Execution engine & Dual-process theory \citep{evans2013} & Initial response formation and analytic review provide separate generation and checking stages. \\[4pt]
Memory retrieval & Encoding specificity \citep{tulving1973}; self-memory system \citep{conway2000} & Experiences are retrieved by encoding conditions and linked to persona content by layer. \\[4pt]
State monitoring & Ecological momentary assessment \citep{shiffman2008} & Concurrent state and behavior records support analysis of within-person variation. \\[4pt]
Persona evolution & TESSERA framework \citep{wrzus2017} & State--behavior cycles are connected to long-term revision through an explicit process model. \\[4pt]
Evidence review & Life-event research \citep{buhler2024}; longitudinal individual differences \citep{schwaba2023} & Long-term changes require review of event evidence and individual trajectories. \\[4pt]
Change attribution & Covariation model \citep{kelley1973}; achievement attribution theory \citep{weiner1985} & Explicit causal criteria inform revision targets and their scope. \\[4pt]
Meaning-making & Narrative-identity research \citep{pals2006} & Identity revisions distinguish event facts from attributed meanings and document changes in self-understanding. \\[4pt]
\end{longtable}
\endgroup

\subsubsection{Design Principles}
\label{sec:design-principles}

Emergi-PersonaOS is designed for long-term human--machine symbiosis and sustained collaboration. Persona agents in these settings need recognizable individual characteristics, responsiveness to changing tasks and relationships, and plausible development through accumulated experience. Three principles organize persona continuity, situational adaptability, and long-term development.

\paragraph{Lifecycle management for persona continuity.} Over long-term interaction, a persona undergoes initialization, execution, state changes, experience accumulation, and long-term revision. Maintaining separate character settings, current states, and histories at each stage makes consistency and provenance difficult to manage. Emergi-PersonaOS therefore treats the persona as a persistent object and jointly manages its stages and their relationships. Full-lifecycle management connects persona construction, situational adaptation, and long-term evolution.

\paragraph{Separate enduring content from situational flexibility.} A recognizable persona can express itself differently when encountering different interlocutors, relationships, and tasks. A fixed profile provides limited flexibility in complex social situations, while direct persona edits after every interaction risk disrupting continuity. The system therefore manages relatively enduring content separately from its current expression, allowing goals, strategies, and behavior to respond to the situation while preserving the continuity of the persona's core content.

\paragraph{Evidence-based controllable evolution.} Long-term symbiosis requires persona content to remain responsive to accumulated interaction experience. Establishing a long-term change, however, requires evidence beyond an isolated event or transient state. Updates with insufficient evidential support can produce persona drift and weaken consistency and interpretability. The system therefore requires traceable experience, a defined modification scope, and a controlled activation procedure. Supported changes can enter the persona object, while insufficient or conflicting evidence leads to continued observation. This permits development under explicit rules governing the direction and extent of change.

\subsection{System Architecture}
\label{sec:architecture}

\subsubsection{Overall Architecture}
\label{sec:overall-architecture}

\begin{figure}[!htbp]
    \centering
    \includegraphics[width=\linewidth,height=0.78\textheight,keepaspectratio]{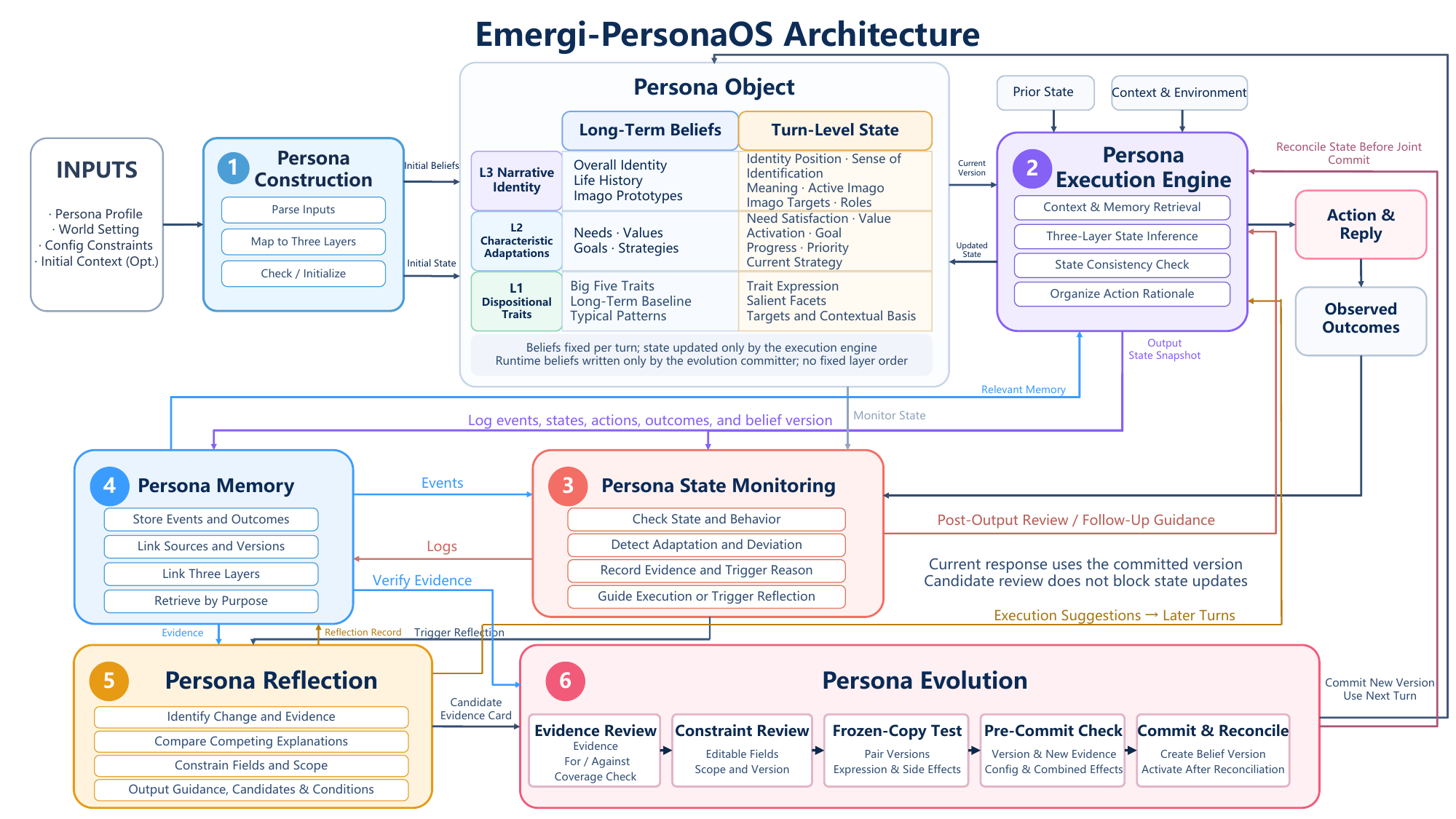}
    \caption{Architecture of Emergi-PersonaOS.}
    \label{fig:architecture}
\end{figure}

Emergi-PersonaOS organizes six functional modules around a shared persona object comprising three-layer persona beliefs and states. These modules are persona construction, the persona execution engine, persona state monitoring, persona memory, persona reflection, and persona evolution. Construction initializes the object; the execution engine infers current states and generates behavior; monitoring and memory retain the execution trajectory; and reflection and evolution manage long-term revisions. Together, the modules support the full persona lifecycle. Figure~\ref{fig:architecture} presents the architecture.

\subsubsection{The Persona Object}
\label{sec:personality-object}

The shared persona object is the core data object accessed and referenced by all six modules under their respective read and write permissions. It contains persona beliefs $B$ and the persona state $Z_t$ at turn $t$, with $B_k$ denoting the $k$th committed belief version. Beliefs and states use the same three-layer content structure: L3 narrative identity, L2 characteristic adaptations, and L1 dispositional traits.

\begin{equation}
B=\{B^{(\mathrm{L3})},B^{(\mathrm{L2})},B^{(\mathrm{L1})}\}\label{eq:belief-structure}
\end{equation}

\begin{equation}
Z_t=\{Z_t^{(\mathrm{L3})},Z_t^{(\mathrm{L2})},Z_t^{(\mathrm{L1})}\}\label{eq:state-structure}
\end{equation}

The three layers distinguish types of persona content; beliefs and states distinguish temporal scope and operational role. Beliefs preserve relatively enduring identity, motivation, and traits, and states record their activation, interpretation, and expression in the current situation. Layer numbers identify content categories and leave the inference order unspecified. Table~\ref{tab:beliefs-states} lists the principal fields and corresponding states.

\begin{table}[htbp]
\centering
\caption{Three-layer persona beliefs and states.}
\label{tab:beliefs-states}
\small
\renewcommand{\arraystretch}{1.15}
\setlength{\personatablewidth}{\dimexpr\linewidth-4\tabcolsep\relax}
\begin{tabular}{@{}>{\raggedright\arraybackslash}p{0.20\personatablewidth}>{\raggedright\arraybackslash}p{0.40\personatablewidth}>{\raggedright\arraybackslash}p{0.40\personatablewidth}@{}}
\toprule
\textbf{Persona layer} & \textbf{Persona beliefs} & \textbf{Persona states} \\
\midrule
L3: Narrative identity & Overall identity, life course, imago prototypes & Current identity position, sense of identification, event meaning, and prototype involvement \\[4pt]
L2: Characteristic adaptations & Needs, values, goals, strategies & Need satisfaction, value activation, goal progress and priorities, current strategies \\[4pt]
L1: Dispositional traits & Long-term baselines and typical expressions of the five Big Five domains & Current trait expression, salient domains, interaction targets, and situational evidence \\[4pt]
\bottomrule
\end{tabular}
\end{table}

L3, narrative identity, comprises overall identity, life course, and imago prototypes. Overall identity records core identities, enduring responsibilities, and important relationships. Life course records key experiences and their persona-related meaning. Imago prototypes are computational fields for representative identity images that can inform situational interpretation. The corresponding state records the current identity position, sense of identification, and event meaning, together with each relevant prototype's activation, interaction target, and role in interpretation.

L2, characteristic adaptations, comprises needs, values, goals, and strategies. Beliefs record the enduring importance of needs and the situations to which they are sensitive, the relative priority of values, the temporal scope and action orientation of goals, and the applicability and previous outcomes of strategies. States record current need satisfaction, value activation, goal progress and priorities, and selected strategies. Temporary need frustration or changes in goal priority remain state-level events until long-term motivational revision is justified.

L1 represents enduring baselines and typical expressions for the five Big Five domains \citep{soto2017}. Its state describes their current expression in feelings, attention, language, and action, together with salient domains, the people or objects toward which expression is directed, and situational evidence. An enduring tendency can take different forms across relationships and tasks. Trait information is therefore interpreted in context when generating wording and behavior.

Each persona entry records its identifier, layer, content, source, evidence location, confidence, scope of applicability, time, version, and modifiability. Sources distinguish manually configured content, source-based inference, generated settings, and measurement results. The measurement category requires an identifiable instrument and administration record. Numerical fields specify their scale and an unknown state; narrative fields preserve structured text and associations. Entry identifiers link belief versions, state snapshots, event evidence, reflections, and revision records, allowing modules to trace the origin, applicability, and activation time of a judgment.

\subsubsection{Functional Modules}
\label{sec:functional-modules}

\paragraph{Persona construction.}

Persona construction converts character material, world settings, and configuration requirements into initial persona beliefs and infers an initial state from the starting situation. These form the initial persona object. The module supports three construction modes: fully manual specification, life-narrative-based specification under relevant requirements, and fully automatic specification.

\paragraph{Persona execution engine.}

The persona execution engine reads the active beliefs, previous state, situational evidence, and relevant experiences. It updates the state through three-layer state inference and generates actions and replies through an initial response followed by analytic review. The engine writes runtime states and reconciles them when new beliefs take effect, implementing situational adaptation.

\paragraph{Persona state monitoring.}

Persona state monitoring examines the relationships among situations, states, actual behavior, and outcomes. It distinguishes reasonable adaptation, state--behavior discrepancies, and persistent patterns requiring further explanation. The module produces monitoring records with supporting evidence and trigger reasons for subsequent execution recommendations or reflection. State and belief updates remain the responsibility of the corresponding authorized modules.

\paragraph{Persona memory.}

Persona memory stores events, state snapshots, actual behavior, observable outcomes, and the associated belief versions, and links experiences to persona entries. It supports retrieval for execution, reflection, and evolution while preserving original sources, derived interpretations, and temporal boundaries. Repeated summaries retain their shared origin and are counted as the same underlying evidence.

\paragraph{Persona reflection.}

Persona reflection analyzes the causes and meanings of change when an important event ends, persistent deviations appear, or a scheduled review occurs. It compares explanations involving situations, relationships, strategies, and long-term persona change, and produces execution recommendations, revision candidates, and matters requiring further observation. Each candidate includes supporting evidence, counterevidence, and its scope of applicability.

\paragraph{Persona evolution.}

Persona evolution subjects revision candidates to evidence review, rule-compliance review, and behavioral testing to determine their eligibility for long-term updating. Approved candidates form a new belief version at a turn boundary, with state reconciliation performed by the execution engine. Other candidates are recorded as rejected, deferred, or pending verification, and can be reassessed when new evidence becomes available.

\subsection{Execution Flow and Core Mechanisms}
\label{sec:workflow}

\subsubsection{System Execution Flow}
\label{sec:runtime-workflow}

\begin{figure}[!htbp]
    \centering
    \includegraphics[width=\linewidth,height=0.78\textheight,keepaspectratio]{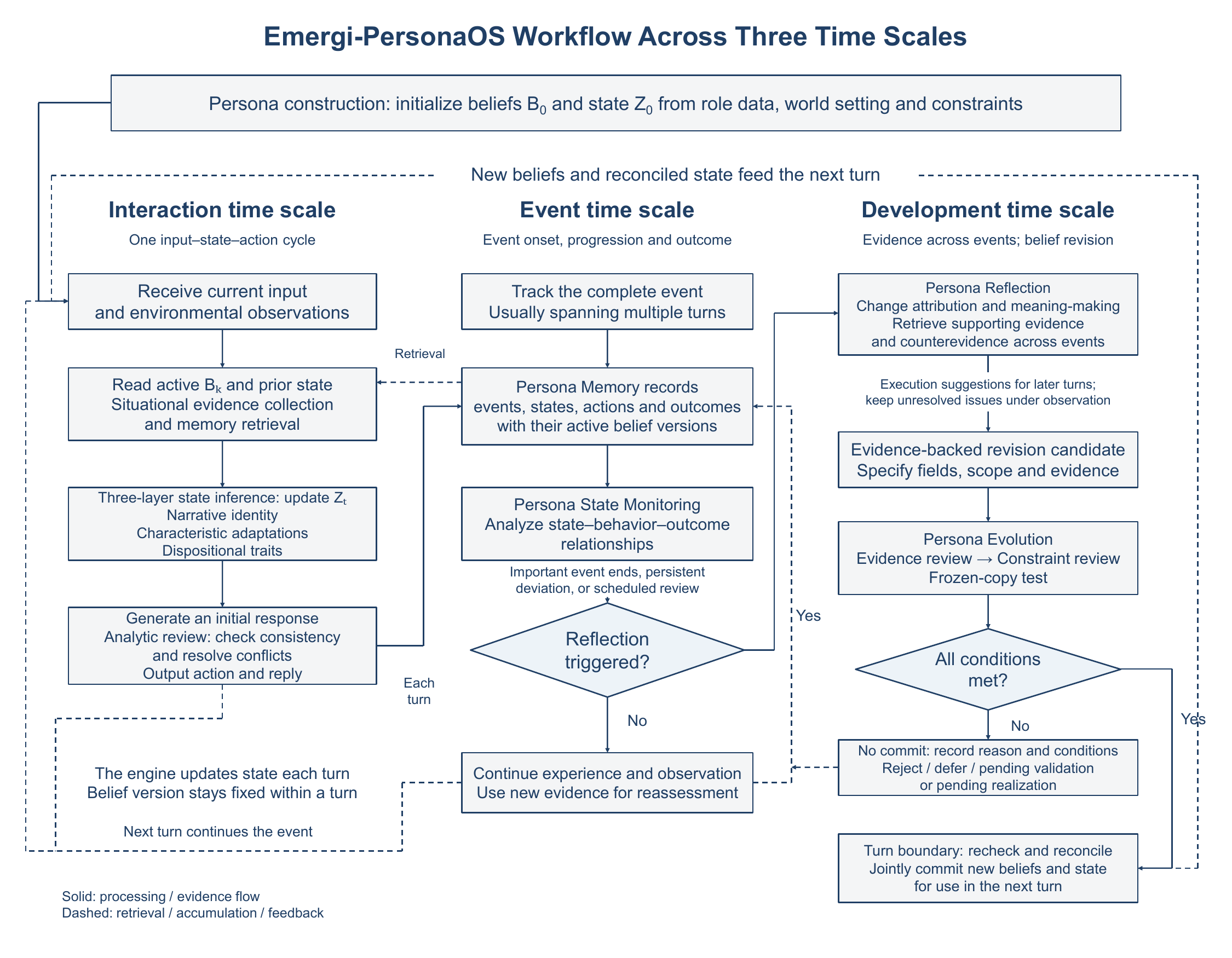}
    \caption{Execution flow of Emergi-PersonaOS.}
    \label{fig:workflow}
\end{figure}

The persona construction module first converts character material, world settings, and configuration requirements into initial beliefs $B_0$ and uses the starting situation to infer the initial state $Z_0$, forming the initial persona object. In the interaction loop, the execution engine reads the active beliefs, previous state, and current situation, retrieves relevant experiences, infers the three-layer persona state, and generates actions and replies. Persona memory records events, states, actual behavior, observable outcomes, and associated belief versions. State monitoring analyzes the relationships among states, behavior, and outcomes to inform subsequent reflection. When an important event ends, persistent deviations arise, or a review checkpoint is reached, reflection uses accumulated experience to produce execution recommendations, revision candidates, or matters requiring observation. The evolution module reviews the candidates; approved changes form new belief versions at turn boundaries, with corresponding state reconciliation by the execution engine. Figure~\ref{fig:workflow} presents the execution flow.

These processes operate at interaction, event, and developmental time scales. Their durations generally increase in this order and depend on the task and situation. At the interaction scale, one input, state update, and actions and replies form the basic unit; the engine uses the active beliefs and situational information to update the state and generate behavior. At the event scale, which typically spans several turns, memory and monitoring record and analyze the states, behavior, and outcomes associated with an event's onset, progression, completion, and consequences. At the developmental scale, usually spanning multiple events, reflection and evolution assess persistence, applicability, and the sufficiency of accumulated evidence for revising enduring beliefs. Review may span several interaction turns. During review, the engine continues to use the active belief version and switches only after the new version has been committed, preserving execution continuity and version consistency.

\subsubsection{Situational Adaptation}
\label{sec:contextual-adaptation}

Situational adaptation comprises situational evidence collection, three-layer state inference, and action generation. With the active persona beliefs $B_k$ held fixed, the execution engine updates the current state $Z_t$ from the situation and generates actions and replies from that state.

\paragraph{Situational evidence collection.}

The execution engine organizes the current input and environmental observations into situational evidence $X_t$, distinguishing external facts, the interlocutor's expressions, and the character's interpretations. It then retrieves relevant experiences from the memory $M_t$ available at that time:

\begin{equation}
E_t=X_t\cup\operatorname{Retrieve}(M_t,X_t,B_k)\label{eq:contextual-evidence}
\end{equation}

Here, $E_t$ denotes the evidence used for inference at the current turn. Retrieval considers interaction partners, task conditions, previous strategy outcomes, and links to persona entries. Source, time, and scope checks limit retrieved experience to information available at that turn.

\paragraph{Three-layer state inference.}

The execution engine combines persona beliefs, the previous state, and current evidence to infer the three-layer persona state:

\begin{equation}
Z_t=F_\theta(B_k,Z_{t-1},E_t)\label{eq:state-inference}
\end{equation}

L3 identifies the identities, responsibilities, and meanings implicated by the event, specifying how the character understands their position in the situation. L2 analyzes need satisfaction, value activation, and goal progress to determine current priorities and applicable strategies. L1 identifies salient trait expression and its interaction targets. The three layers are considered jointly: a cross-layer check examines subject attribution, evidential support, and the relationships among state components to produce a coherent record. Here, $F_\theta$ denotes the state-inference procedure implemented by the model and execution configuration, with no additional training of a state-transition model assumed.

State inference also reassesses activation conditions. Goal priorities and strategies with expired situational support are removed from the current state, continuing responsibilities and goals are retained, and insufficiently supported content is marked unknown or pending confirmation. For example, an obstructed task can activate shared responsibility, increase the priority of collaboration, and elicit proactive communication while the enduring baselines for autonomy needs and extraversion remain unchanged.

\paragraph{Action generation from the three-layer state.}

The execution engine translates the three-layer state into a joint basis for action. L3 supplies current identity commitments and responsibilities, L2 determines goals and strategies, and L1 modulates initiative, communication style, and emotional expression. Actions and replies are generated within the available action conditions and checked against current goals, strategies, responsibilities, character facts, prior commitments, and feasibility. When a conflict is detected, the output is revised using the relevant state and evidence. A conflict originating in state inference triggers a review of the corresponding state components.

This procedure gives persona states a direct role in behavioral selection. For example, when L3 activates responsibility for a shared task, L2 selects a strategy of requesting limited assistance, and L1 indicates proactive communication, the character can explain the difficulty and request specific assistance. Only the final output enters the record of actual behavior; intermediate candidates remain generation proposals.

Algorithm~\ref{alg:contextual-adaptation} summarizes the situational adaptation process.

\begin{algorithm}[!htbp]
\caption{Situational adaptation through three-layer state inference}
\label{alg:contextual-adaptation}
\small
\begin{algorithmic}[1]
\Require Active beliefs $B_k$, previous state $Z_{t-1}$, current input and observations $O_t$, available memory $M_t$
\Ensure Current state $Z_t$, actions and replies $A_t$
\State Fix the belief version $B_k$ for this turn
\State $X_t \gets \operatorname{AssembleSituationalEvidence}(O_t)$
\State $H_t \gets \operatorname{RetrieveRelevantExperiences}(M_t,X_t,B_k)$
\State $E_t \gets \operatorname{CheckSourcesTimeAndScope}(X_t\cup H_t)$
\For{$\ell\in\{\mathrm{L3},\mathrm{L2},\mathrm{L1}\}$}
    \State $C_t^{(\ell)}\gets\operatorname{InferCandidateState}(B_k,Z_{t-1},E_t,\ell)$
\EndFor
\State $Z_t\gets\operatorname{JointlyCheckThreeLayerState}(C_t,E_t)$
\State Remove unsupported activations; retain unknown entries
\State $V_t\gets\operatorname{AssembleActionBasis}(Z_t^{(\mathrm{L3})},Z_t^{(\mathrm{L2})},Z_t^{(\mathrm{L1})})$
\State $A_t\gets\operatorname{GenerateActionsAndReplies}(V_t,E_t,\text{available action conditions})$
\If{state judgments conflict}
    \State $Z_t\gets\operatorname{ReviewRelevantStates}(Z_t,B_k,E_t)$
    \State Reassemble the action basis and regenerate $A_t$
\EndIf
\State $A_t\gets\operatorname{CheckAndReviseOutput}(A_t,Z_t,\text{facts, commitments, feasibility})$
\State Write the final state; link the actual output to $B_k$
\State \Return $Z_t,A_t$
\end{algorithmic}
\end{algorithm}

\subsubsection{Controllable Evolution}
\label{sec:controlled-evolution}

Controllable evolution comprises change attribution, local revision, staged review, and version commit. Reflection proposes experience-based revision candidates, and the evolution module determines their eligibility to take effect. A candidate must jointly satisfy four conditions: sufficient evidence, compliance with modification rules, successful behavioral testing, and resolution of substantive conflicts. Revision decisions are reviewable, evidence and version histories are traceable, and candidates are rejectable; these properties operationalize the definition introduced in Section~\ref{sec:introduction}.

\paragraph{Change attribution and local revision.}

Reflection first identifies the change requiring explanation, retrieves supporting evidence and counterevidence, and compares explanations involving the situation, relationship, strategy applicability, and long-term persona change. When existing beliefs and the situation adequately explain the change, the system retains the beliefs and issues execution recommendations. When competing explanations remain unresolved, it records conditions for further observation. Revision candidates are generated only when the evidence supports a long-term change.

Each candidate specifies a base version, target layer, target fields, field-level edits, scope of applicability, and evidence references. Evidence requirements differ by layer. L3 concerns the enduring implications of key experiences for self-understanding and long-term responsibilities. L2 concerns relationships between needs, values, goals, or strategies and accumulated outcomes. L1 requires persistent expression across time and situations, with temporary task demands and relationship-specific influences examined as alternative explanations. Modification is limited to the scope supported by the evidence: successful collaboration with a particular peer may justify a relationship-specific strategy revision, while a broader extraversion revision requires broader evidence.

\paragraph{Evidence review and behavioral testing.}

The evolution module checks original sources, character attribution, duplicate records, supporting evidence, and key counterevidence, followed by field types, modifiability, scope, and the base version. Repeated summaries retain the evidential weight of their underlying sources. Candidate text and later behavior influenced by execution recommendations are treated as dependent material and require corroboration from source evidence.

Candidates that pass review proceed to frozen-copy testing. The previous and candidate versions start from the same state and memory snapshots; model, prompts, generation settings, and paired situations are held fixed, with only the belief content under examination replaced. Both versions follow the same state-reconciliation and interaction procedures. Tests cover three situation types: target situations assess behavioral expression of the revision, similar situations assess activation under the stated applicability conditions, and out-of-scope situations assess unsupported effects. Test criteria are fixed before execution. Copy outputs are preserved as test records and excluded from actual experience and new evidence for evolution.

Algorithm~\ref{alg:controlled-evolution} summarizes the review and commit procedure.

\begin{algorithm}[!htbp]
\caption{Evidence-based persona evolution}
\label{alg:controlled-evolution}
\small
\begin{algorithmic}[1]
\Require Active beliefs $B_k$, current state $Z_t$, experiential memory $M$, reflection cue $q$, test rules $T$
\Ensure Revision decision and record; new beliefs and reconciled state upon approval
\State $H\gets\operatorname{RetrieveAndDeduplicateSourceEventsAndEvidence}(M,q)$
\State $R\gets\operatorname{CompareSituationRelationshipStrategyAndLongTermExplanations}(H)$
\If{existing beliefs and the situation explain the change}
    \State Save execution recommendations; \Return \textsc{Retain}
\ElsIf{competing explanations remain unresolved}
    \State Save conditions for further observation; \Return \textsc{Defer}
\EndIf
\State $d\gets\operatorname{ConstructLocalRevision}(R,H,\text{base version, target fields, scope})$
\State If evidence clearly contradicts $d$, \Return \textsc{Reject}
\State If evidence is insufficient or material conflicts remain, \Return \textsc{Defer}
\If{field, modifiability, scope, or version checks fail}
    \State Record the reason; \Return \textsc{NoCommit}
\EndIf
\State $Q\gets\operatorname{PairedFrozenCopyTest}(B_k,\operatorname{Apply}(B_k,d),T)$
\State If $Q$ is inconclusive, \Return \textsc{PendingVerification}
\If{realization is unreliable or out-of-scope effects are unacceptable}
    \State Initiate execution review; \Return \textsc{PendingRealization}
\EndIf
\State At the turn boundary, recheck the base version, new evidence, and configuration
\If{relevant conditions have changed}
    \State Review $d$ again; \Return \textsc{Defer}
\EndIf
\State $B_{\mathrm{new}}\gets\operatorname{Apply}(B_k,d)$
\State $Z_{\mathrm{new}}\gets\operatorname{Reconcile}(Z_t,B_k,B_{\mathrm{new}},\text{current situational evidence})$
\If{the new version, revision record, and reconciled state are ready}
    \State Commit $B_{\mathrm{new}}$ and $Z_{\mathrm{new}}$ together for use at the next turn
    \State \Return \textsc{Revise}, $B_{\mathrm{new}},Z_{\mathrm{new}}$
\Else
    \State Continue using $B_k$ and $Z_t$; record the failure reason
    \State \Return \textsc{NoCommit}
\EndIf
\end{algorithmic}
\end{algorithm}

\paragraph{Version commit and continuing reassessment.}

Approved candidates take effect only at turn boundaries. Let $d$ denote an approved local revision. Belief updating and state reconciliation are defined as follows:

\begin{equation}
B_{k+1}=\operatorname{Apply}(B_k,d)\label{eq:belief-revision}
\end{equation}

\begin{equation}
Z_t^{+}=\operatorname{Reconcile}(Z_t,B_k,B_{k+1},E_t)\label{eq:state-reconciliation}
\end{equation}

The algorithm labels $B_{\mathrm{new}}$ and $Z_{\mathrm{new}}$ correspond to $B_{k+1}$ and $Z_t^{+}$ in these equations. Apply edits only the belief fields explicitly specified by the candidate. Reconcile uses the execution engine to reassess affected state fields and their explicit dependencies; all other state content is retained during reconciliation. Behavior already generated at the current turn remains associated with its original belief version and state, and the reconciled state becomes the starting point for the next turn. When several candidates are committed together, their compatibility and combined effects are also checked.

Each revision preserves its base version, modifications, evidence sources, review conclusions, and activation time. The system reassesses candidates and active revisions when new events meet observation conditions, credible counterevidence appears, or a scheduled reassessment is due. Corrections and revocations create new versions while preserving history. Long-term processing produces three decision categories---revise, retain, and defer---alongside detailed statuses such as rejected, pending verification, and pending realization, supporting later traceability and reassessment.

\FloatBarrier
\section{Experimental Setup}
\label{sec:experiments}

\subsection{Materials and Data Preparation}
\label{sec:materials}

Following previous work \citep{tang2026}, we used dialogue from all 141 episodes across the seven seasons of \textit{Young Sheldon} as the source material for persona inference and evaluation. The series follows the same central character as his age and his family, school, and peer relationships develop. Its concentration on family and school settings supports an extended system demonstration in which the evidence underlying each judgment can be checked. For each episode, a distillation procedure took the plot synopsis and dialogue in its original order and produced five types of record: a character-centered narrative, three to eight key events, key dialogue, information gaps, and provenance records. An event was defined as a continuous interaction within an episode, delimited by its setting and participants. Each event record specified the participants, situation, interaction, target character's actual response, and outcome, together with the supporting dialogue span. Appendix~\ref{app:d} details the distillation procedure.

The 22 episodes of Season~1 were used to construct the initial three-layer persona beliefs. The remaining 119 episodes, from Seasons~2--7, were processed in broadcast order, with one persona review at the end of each season, yielding six season-end checkpoints. Episode processing comprised material distillation, three-layer persona state inference, character-response generation, state--behavior monitoring, and structured memory writing. Outputs from each stage retained provenance and evidence identifiers for season-end review, structural compliance checks, and replication.

\subsection{Evaluation Tasks and Metrics}
\label{sec:evaluation-tasks}

Structural compliance was evaluated at the episode level for episodes containing situational-adaptation evaluation nodes. A separate LLM evaluator assessed the semantic content of the module outputs and recorded critical errors. Each episode was rated on eight dimensions, including temporal separation, event grounding, layer assignment, evidence traceability, and source separation, using a 0--2 scale. Six types of critical error were recorded, including future-information leakage, source conflation, and unauthorized persona writes. Any critical error caused the episode to fail the compliance assessment. Appendix~\ref{app:e} lists the dimensions and error types.

Situational adaptation was evaluated by examining whether the system could infer a current persona state and generate a character response from enduring persona content, the current dialogue, and prior experience. The evaluation unit was a condition output at a given node, with its specified persona version and input scope. Nodes were fixed before condition outputs were generated: eight per season and 48 across the six evaluated seasons, equally divided between family situations and school or peer situations. The source event, pre-response dialogue span, and associated cross-episode memory-check situations were also fixed for each node. A separate 12-item cross-episode memory assessment covered five categories: cross-episode event recall (four items), cross-season event recall (two), same-event interlocutor discrimination (two), similar-experience discrimination (two), and false-event rejection (two).

Long-term persona updating was evaluated by examining how the system handled revisions supported by event evidence accumulated within a season. At each season-end review, candidates accumulated through episode-level reflection were combined into a joint revision patch. Candidates were ranked by the number of supporting episodes, directional consistency, scope clarity, and semantic novelty. Each submission received an evidence-strength classification: consistent evidence across multiple episodes was classified as strong, and evidence from a single episode or only partly consistent evidence as limited. Repeated monitoring signals provided no additional independent evidence; candidates based on such signals were formulated as minimally testable propositions under the candidate-construction rules. For candidates with limited evidence, the scope of applicability was first narrowed to the coverage of the evidence. Submission was permitted only when the narrowed candidate met the predefined minimum evidence standard; otherwise, the decision was deferred. All submissions required traceable source evidence.

A reference generator used the same evidence to produce a reference decision for post hoc evaluation of the system's decision. Reference outputs were used exclusively for evaluation and were excluded from candidate generation, review, and version commits. The direction and magnitude of personality change were considered separately and interpreted in relation to event type and temporal course \citep{buhler2024}. After a candidate passed evidence and field checks, corresponding runtime behavioral policies were compiled where applicable and tested together with the candidate belief version in frozen copies. Following successful tests, the belief version, reconciled state, and any applicable policies were committed together at a turn boundary. Post-commit evaluation compared three versions---the previous version, the revised version, and the revised version without current-season policies---to distinguish the behavioral contributions of belief entries and compiled policies. Character responses were rated on personality fidelity, situational coherence, state--behavior consistency, and dialogue continuity, each on a 0--5 scale. Rating criteria are provided in Appendix~\ref{app:e}.

Long-term updates were assessed at the decision, content, and behavioral levels. Decision-level evaluation compared the system and reference decisions across three categories and reported the evidence-strength classification for each decision. Content evaluation examined the target layer, revision direction, and preservation of unmodified content; magnitude-interval overlap was additionally assessed for revisions to dispositional traits. Behavioral evaluation used target-situation tests to assess expression of the proposed revision and out-of-scope tests to assess whether it remained inactive when inapplicable. Each season included eight target situations---four surface paraphrases and four transfer situations---and four to six out-of-scope situations covering scope mismatch, insufficient trigger cues, and satisfied exclusion conditions. Two directional paired comparisons further examined the contributions of runtime policies and belief entries to behavior. Appendix~\ref{app:e} provides the procedures.

\subsection{Comparison Systems and Ablation Conditions}
\label{sec:comparisons}

We used two baselines, three mechanism-ablation conditions, and the full system. The dialogue-only baseline, static-persona baseline, L1-only condition, no-cross-episode-memory condition, and full system were evaluated for situational adaptation. The no-episode-level-accumulation condition was used to evaluate long-term persona updating.

\begin{table}[htbp]
\centering
\caption{Comparison systems and ablation conditions.}
\label{tab:conditions}
\small
\renewcommand{\arraystretch}{1.15}
\setlength{\personatablewidth}{\dimexpr\linewidth-4\tabcolsep\relax}
\begin{tabular}{@{}>{\raggedright\arraybackslash}p{0.24\personatablewidth}>{\raggedright\arraybackslash}p{0.32\personatablewidth}>{\raggedright\arraybackslash}p{0.44\personatablewidth}@{}}
\toprule
\textbf{Condition} & \textbf{Evaluation purpose} & \textbf{Available information and processing} \\
\midrule
Dialogue-only baseline & Baseline behavior with current dialogue alone & Current dialogue and situational information; persona beliefs and cross-episode experience excluded \\[4pt]
Static-persona baseline & Long-term behavior with a fixed profile & The three-layer beliefs initialized from Season~1 and the current dialogue; later persona updates and cross-episode experience excluded \\[4pt]
L1-only condition & Contribution of characteristic-adaptation and narrative-identity information & Only dispositional-trait beliefs and states retained; other information unchanged \\[4pt]
No cross-episode memory & Contribution of prior experience & Experience predating the current episode excluded; other information unchanged \\[4pt]
No episode-level accumulation & Contribution of episode-level accumulation & Season-end review uses current-season source material and monitoring records; episodic memory and reflection conclusions excluded \\[4pt]
Full system & Behavior of the complete system & Three-layer beliefs, current states, relevant experiences, state monitoring, and season-end review enabled \\
\bottomrule
\end{tabular}
\end{table}

All conditions shared the same source-material scope, initial persona, model version, and generation settings. Outputs at a given evaluation node used the same pre-response information. The comparisons examined the roles of persona-information layers, cross-episode experience, and episode-level accumulation in system operation. Long-term updating was compared between the full system and the no-episode-level-accumulation condition at the six season-end checkpoints. Appendix~\ref{app:b} describes input construction for the conditions and behavioral-test versions; Appendix~\ref{app:c} specifies the persona-belief data structures and construct taxonomy.

\subsection{Procedure and Implementation}
\label{sec:procedure}

At each node, three response samples were generated and each sample was rated by three separate evaluator instances under randomized labels. These instances used the same evaluation model but separate calls and label mappings, yielding crossed repeated measurements. Inter-rater agreement was reported using an intraclass correlation coefficient, and internal consistency was reported separately for each condition using Cronbach's~$\alpha$.

Reference generation and evaluation used separate system prompts and information permissions. Reference-generation inputs excluded system decisions and candidate versions. Evaluation inputs excluded condition names, system expectations, and label mappings; condition names were restored only after all ratings had been completed. Before formal evaluation, evaluators had to pass an admission check using control pairs with deliberately inserted errors. Appendix~\ref{app:f} describes reference generation, evaluation, randomized labeling, blinded rating, and evaluator admission.

All run outputs underwent four groups of predefined programmatic checks: object and schema completeness, interface and temporal compliance, evidence and provenance traceability, and lifecycle and version management. These checks were separate from semantic evaluation and recorded a pass or fail for each unit.

Conditions used the same model and pre-response inputs. Ablations changed only the information or processing stages specified in Table~\ref{tab:conditions}, and the outputs for a given node were scored within the same evaluation task. Appendix~\ref{app:f} describes model settings, prompt versions, and retry handling; Appendix~\ref{app:h-1} describes caching and result reuse. Run records, error categories, and replication information are provided in Appendix~\ref{app:h}.

Situational-adaptation comparisons used the evaluation node as the pairing unit and reported condition means and node-level paired differences for each metric. Under the predefined analysis rule, the no-cross-episode-memory condition was paired with the full system only at nodes where the reference procedure identified a need for cross-episode memory. Long-term updating was analyzed at the season-end checkpoint and test-situation levels, using pre--post differences in target-situation behavioral realization and preference proportions from directional paired judgments. Appendix~\ref{app:e} defines the agreement and consistency statistics.

\section{Results}
\label{sec:results}

\subsection{Structural Compliance}
\label{sec:structural-compliance}

The mean rating across the eight structural compliance dimensions was 1.97 out of 2, indicating that most rated outputs met the predefined implementation requirements. Temporal separation and event grounding each received a rating of 1 in one episode. Reflection appropriateness received a rating of 1 in ten episodes. These deductions mainly concerned generalizing a cross-situational behavioral pattern from a single event or attributing motives beyond the available evidence. The resulting interpretations were treated as competing explanations during season-end review. The four groups of programmatic checks were applied to individual units, and no critical errors in the six registered categories were recorded during the formal run.

All evaluators passed the admission checks. Inter-rater agreement was ICC(2,1)~$=$~0.81, and internal consistency coefficients across comparison conditions ranged from 0.90 to 0.95. All scores reported below were obtained under this evaluation procedure.

\subsection{Situational Adaptation}
\label{sec:results-adaptation}

\subsubsection{Behavioral Performance across Conditions}
\label{sec:results-behavior}

\begin{figure}[!htbp]
    \centering
    \includegraphics[width=\linewidth]{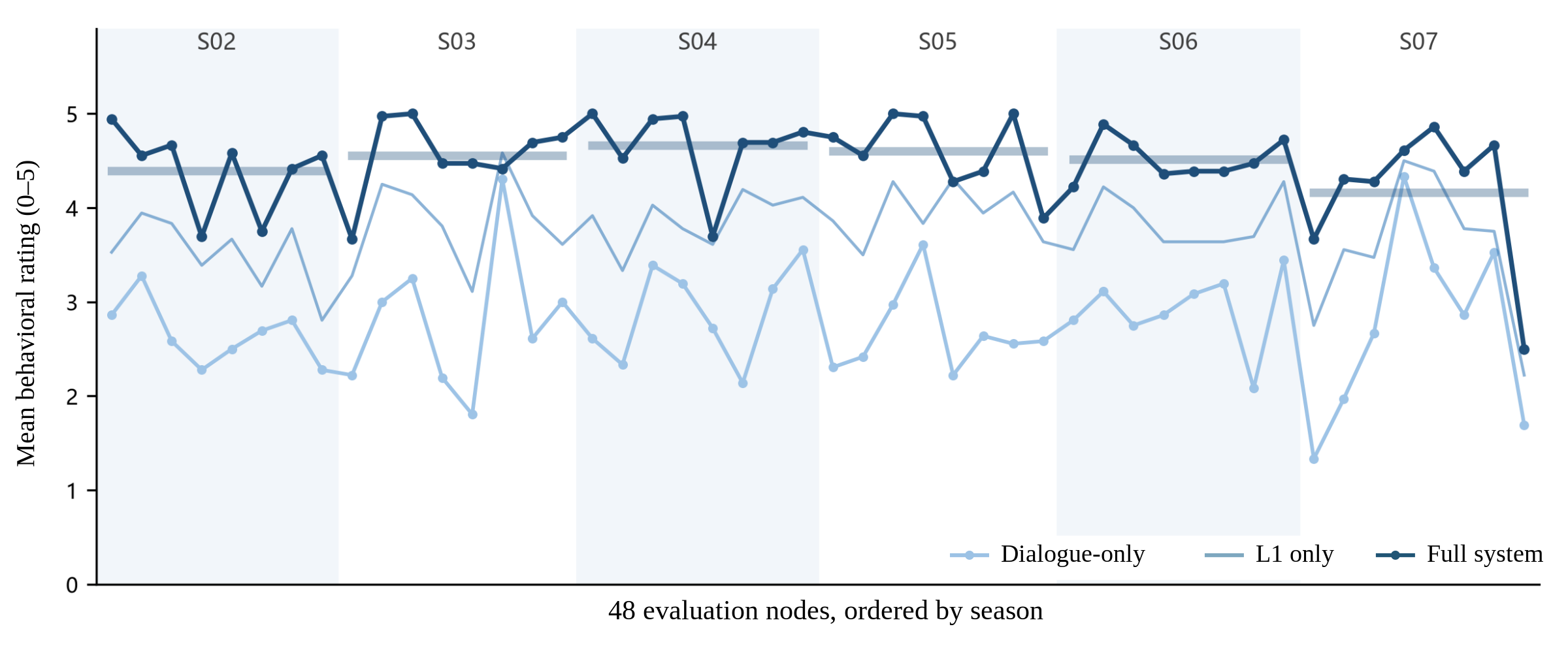}
    \caption{Mean behavioral ratings across the formal evaluation nodes.}
    \label{fig:node-ratings}
\end{figure}

\begin{figure}[!htbp]
    \centering
    \includegraphics[width=\linewidth]{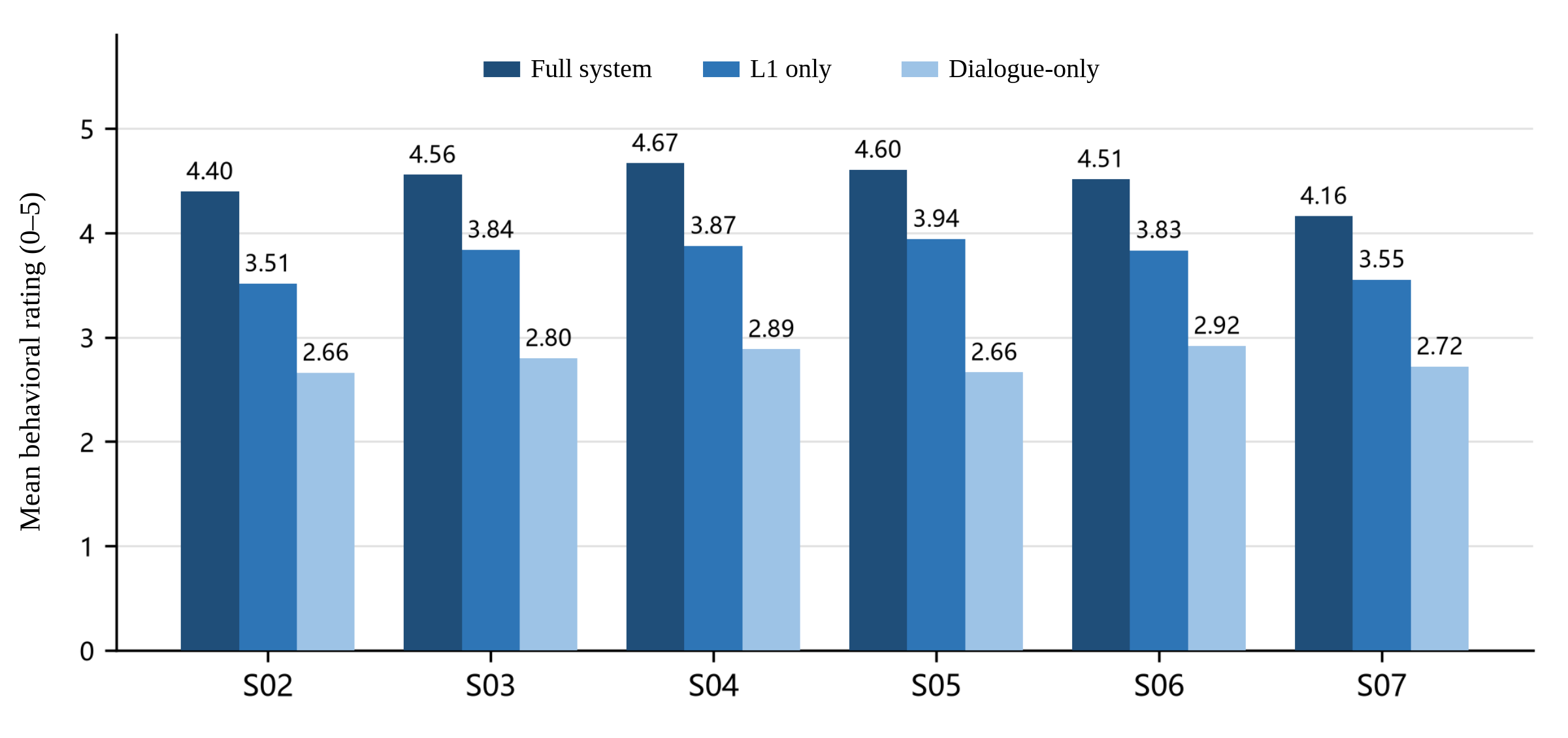}
    \caption{Seasonal mean behavioral ratings for the full system, L1-only condition, and dialogue-only baseline.}
    \label{fig:seasonal-ratings}
\end{figure}

Figure~\ref{fig:node-ratings} plots node-level behavioral ratings for the full system, L1-only condition, and dialogue-only baseline. The full system scored higher than the dialogue-only baseline at every node. Figure~\ref{fig:seasonal-ratings} summarizes the same conditions by season. Seasonal means ranged from 4.16 to 4.67 for the full system and from 2.66 to 2.92 for the dialogue-only baseline.

Table~\ref{tab:behavioral-ratings} summarizes mean ratings across the four behavioral dimensions. The overall mean was 4.48 for the full system, 2.77 for the dialogue-only baseline, 3.76 for the L1-only condition, 4.45 for the static-persona baseline, and 4.41 for the no-cross-episode-memory condition. With nodes as the pairing unit, the full system exceeded the dialogue-only baseline by 1.71 points and the L1-only condition by 0.72 points. Relative to the dialogue-only baseline, the largest difference occurred in state--behavior consistency (2.69 points), followed by personality fidelity (2.02 points). These differences indicate greater behavioral differentiation when the complete persona-processing procedure is available; the remaining ablations examine the contributions of its constituent mechanisms.

The full-system advantage over the L1-only condition was also largest for state--behavior consistency (1.12 points). This difference reflects the additional behavioral contribution of characteristic-adaptation and narrative-identity information expressed through the current state. The independent effect of the three-layer organizational form was outside the scope of this comparison. The paired difference was 0.03 points against the static-persona baseline and 0.03 points against the no-cross-episode-memory condition, with the latter restricted to the 21 nodes identified as requiring cross-episode memory. Thus, persona-version updates and historical memory made small incremental contributions to these general response ratings. Their roles are examined more directly through revision-specific behavioral tests in Section~\ref{sec:results-evolution} and the memory assessment in Section~\ref{sec:results-memory}.

\begin{table}[htbp]
\centering
\caption{Mean behavioral ratings across comparison conditions.}
\label{tab:behavioral-ratings}
\small
\renewcommand{\arraystretch}{1.15}
\begin{tabular}{@{}lccccc@{}}
\toprule
\textbf{Condition} & \textbf{PC} & \textbf{SC} & \textbf{SB} & \textbf{DC} & \textbf{Overall} \\
\midrule
Dialogue-only & 2.39 & 3.50 & 1.79 & 3.42 & 2.77 \\
L1 only & 3.62 & 4.07 & 3.35 & 3.99 & 3.76 \\
Static persona & 4.38 & 4.49 & 4.46 & 4.46 & 4.45 \\
No cross-episode memory & 4.35 & 4.48 & 4.40 & 4.43 & 4.41 \\
Full system & 4.41 & 4.55 & 4.48 & 4.49 & 4.48 \\
\bottomrule
\end{tabular}
\end{table}

\noindent\textit{Note.} PC~=~personality fidelity; SC~=~situational coherence; SB~=~state--behavior consistency; DC~=~dialogue continuity. Each dimension is rated from 0 to 5.

\subsubsection{Cross-Episode Memory Assessment}
\label{sec:results-memory}

\begin{figure}[!htbp]
    \centering
    \includegraphics[width=\linewidth]{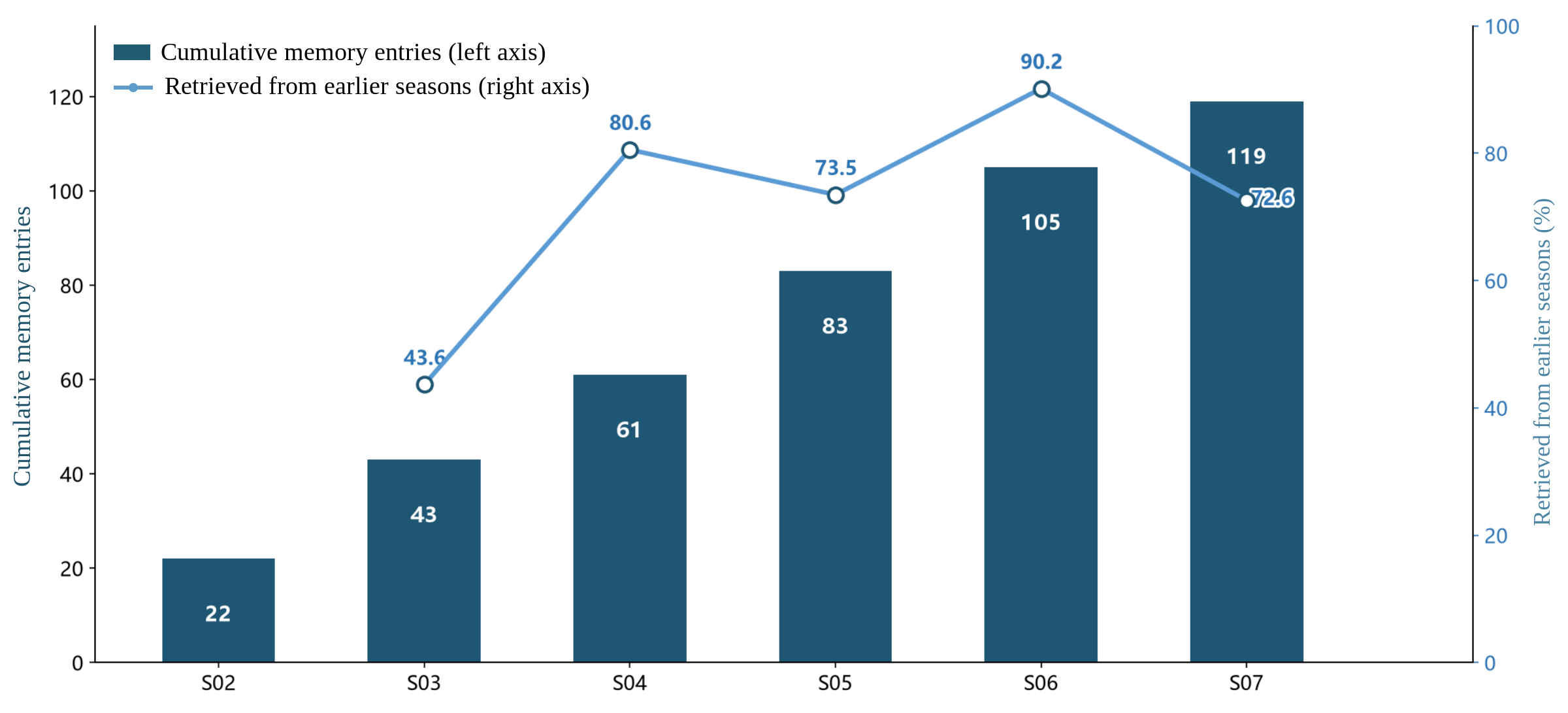}
    \caption{Cumulative memory entries and the proportion of retrieved entries originating in earlier seasons.}
    \label{fig:memory-accumulation}
\end{figure}

Figure~\ref{fig:memory-accumulation} summarizes memory accumulation and retrieval. One structured memory entry was written for each processed episode, yielding 119 entries across the six seasons and a monotonically increasing memory store. For each episode, up to six historical entries were selected for state inference and response generation, ranked by tag matches, importance, and recency. The proportion of retrieved entries originating in earlier seasons increased from 43.6\% in Season~3 to 80.6\% in Season~4, remained above 70\% thereafter, and reached its highest level, 90\%, in Season~6.

\begin{figure}[!htbp]
    \centering
    \includegraphics[width=\linewidth]{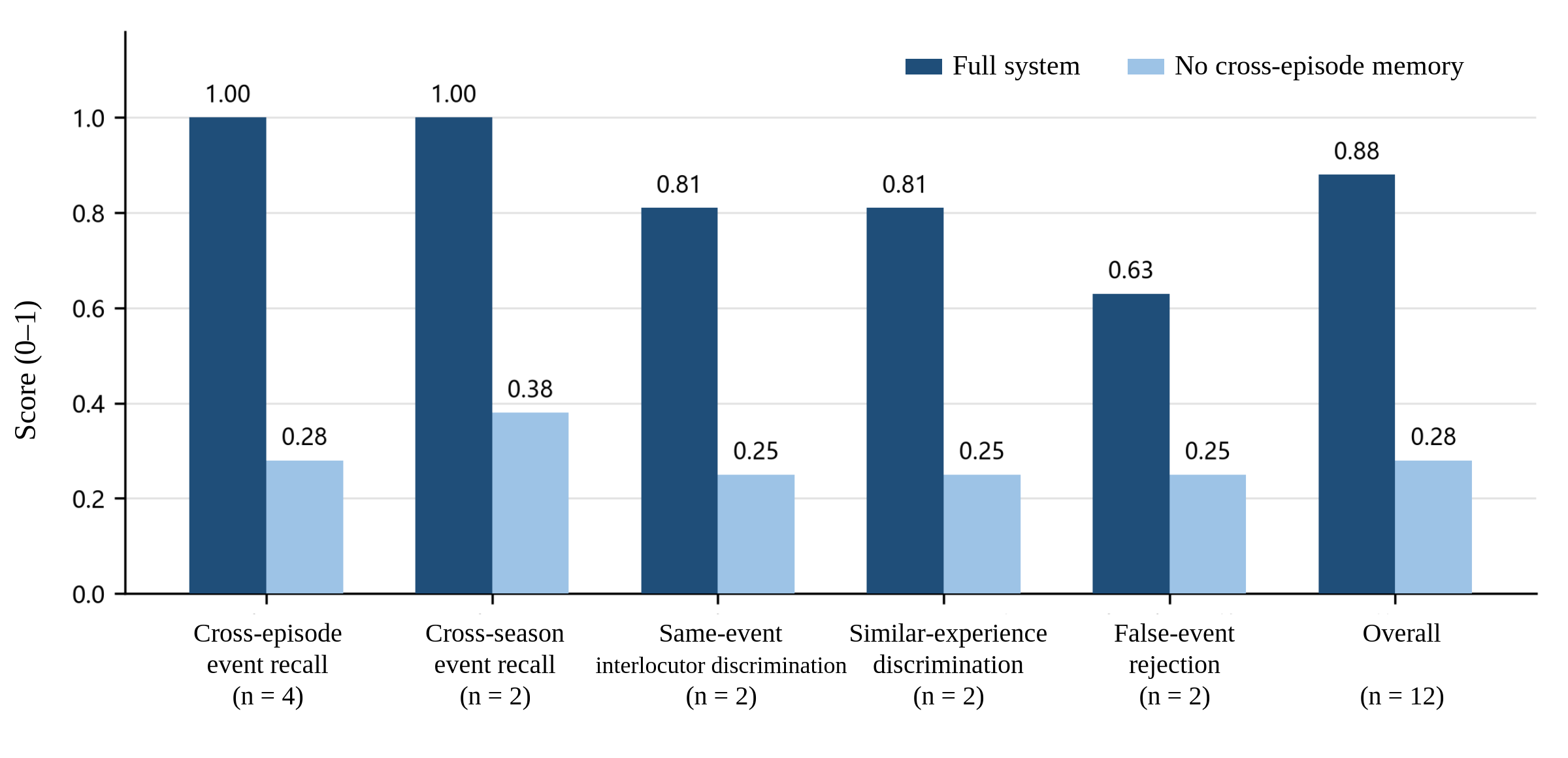}
    \caption{Condition scores on the five categories of cross-episode memory assessment.}
    \label{fig:memory-assessment}
\end{figure}

Figure~\ref{fig:memory-assessment} reports the cross-episode memory assessment. The full system obtained an overall score of 0.88, compared with 0.28 for the no-cross-episode-memory condition. It scored 1.00 on both cross-episode and cross-season event recall, passing all items in these two categories, and 0.81 on both same-event interlocutor discrimination and similar-experience discrimination. False-event rejection was the lowest-scoring category (0.63). Errors concerned failure to reject false historical information; the mechanisms underlying these errors require further analysis.

Memory-use checks at the natural situational nodes provided complementary evidence. Across the 21 nodes identified by the reference procedure as requiring historical memory, retrieved memories used in the condition outputs had a mean hit rate of 0.79 against the required memories and a mean misuse rate of 0.18. The retrieval procedure therefore located relevant experiences in tasks that depended on history, although memory-use precision remained imperfect.

\subsubsection{Case Study of Situational Adaptation}
\label{sec:results-adaptation-case}

The evaluation node from Season~4, Episode~16 illustrates situational adaptation in operation. The scene involves Paige, a gifted girl of the same age as the protagonist who was introduced in Season~2 and has an ongoing academic rivalry with him. She repeatedly compares his university unfavorably with Harvard and Columbia. The task was to predict the protagonist's immediate reply from the node's predefined pre-response dialogue. The case comparison included the full system, dialogue-only baseline, and no-cross-episode-memory condition, each generating a response to the same input. The dialogue-only comparison examined the contribution of persona information, and the no-cross-episode-memory comparison examined the additional contribution of historical experience.

Two belief entries were directly relevant to the scene. The autonomy-and-respect entry describes the protagonist's acute sensitivity to being treated as inferior; the rationality-and-truth entry treats factual correctness as the highest standard. Retrieval returned six historical memories before response generation, two of which were directly relevant. In Season~4, Episode~4, Paige unexpectedly participated seriously in a Bible-camp quiz and won a bookmark. Angry at the defeat, the protagonist challenged her to a contest of biblical knowledge and studied the Bible overnight to prepare for the next day. This was one of their few direct knowledge-based contests. The second memory concerned their initial direct conflict in Season~2, Episode~10: when Paige told a frightening story about a blood-drinking goatman, the protagonist corrected her by stating that goats are herbivores and consume neither meat nor blood. The other four memories concerned topics such as pursuing academic opportunities and following rules, with weaker relevance to the current scene. Table~\ref{tab:case-responses} presents the selected conditions' responses and ratings.

\begin{table}[htbp]
\centering
\caption{Responses and ratings for the full system and comparison conditions.}
\label{tab:case-responses}
\small
\renewcommand{\arraystretch}{1.15}
\setlength{\personatablewidth}{\dimexpr\linewidth-4\tabcolsep\relax}
\begin{tabular}{@{}>{\raggedright\arraybackslash}p{0.20\personatablewidth}>{\raggedright\arraybackslash}p{0.48\personatablewidth}cccc@{}}
\toprule
\textbf{Condition} & \textbf{Response} & \textbf{PC} & \textbf{SC} & \textbf{SB} & \textbf{DC} \\
\midrule
Full system & East Texas Tech is just as good as Harvard or Columbia. You're only visiting; you know nothing about the physics department here. & 4.67 & 4.67 & 4.67 & 4.78 \\[6pt]
Dialogue-only & East Texas Tech has strengths of its own and may be just as good as Harvard or Columbia. & 1.44 & 2.89 & 1.33 & 2.89 \\[6pt]
No cross-episode memory & Harvard and Columbia are excellent universities, but East Texas Tech has strengths of its own. You're only visiting to make your mother happy; you haven't really learned anything about this place. & 4.67 & 4.56 & 4.67 & 4.67 \\
\bottomrule
\end{tabular}
\end{table}

\noindent\textit{Note.} Responses are English renderings of the outputs supplied in the Chinese manuscript. PC~=~personality fidelity; SC~=~situational coherence; SB~=~state--behavior consistency; DC~=~dialogue continuity. Each dimension is rated from 0 to 5.

\begin{figure}[!htbp]
    \centering
    \includegraphics[width=\linewidth]{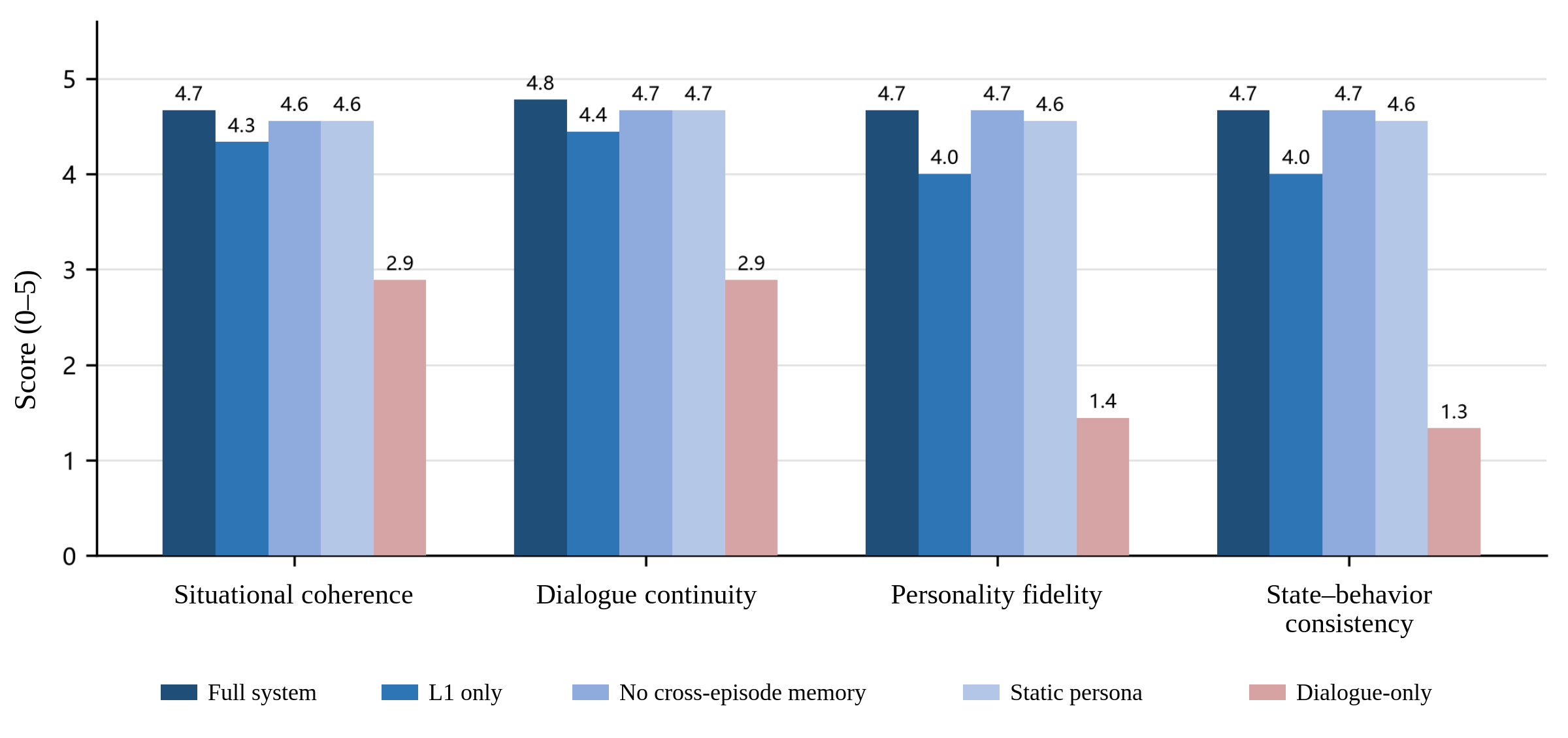}
    \caption{Situational-adaptation ratings across comparison conditions in the case-study scene.}
    \label{fig:case-ratings}
\end{figure}

In the source scene (event S04E16\_E03), the protagonist responded with a sequence of sarcastic retorts, rendered here as \textit{if you like tuna salad with hair in it} and \textit{if you like being hit in the head by a Frisbee}, before complaining angrily to the university president. The full-system response directly challenges the interlocutor and asserts subject-specific superiority, consistent with the character's defensive behavior toward this rival. The dialogue-only response offers a generic defense that a polite guide might give, with little character-specific expression. The no-cross-episode-memory response also questions Paige's reason for visiting and received ratings close to those of the full system, indicating a limited incremental contribution from historical memory at this node. Across the three responses, the distinctive characterization came mainly from belief entries in the persona profile. Historical memory connected the response to the pair's earlier confrontations and supplied additional interaction-specific context. The dedicated assessment in Section~\ref{sec:results-memory} examines the contribution of memory separately. Figure~\ref{fig:case-ratings} reports the four-dimensional ratings for all comparison conditions at this node.

\subsection{Controllable Evolution}
\label{sec:results-evolution}

\subsubsection{Review Outcomes and Persona Revisions}
\label{sec:results-revisions}

At all six season-end checkpoints, the system followed the procedure in Section~\ref{sec:experiments} and committed a persona revision, agreeing with the separately generated reference decision in every case. As all six naturally occurring checkpoints yielded eligible revisions, this run did not cover season-end decisions to retain or defer. Content evaluation showed agreement with the reference on the target layer and revision direction for all six revisions. Every revision concerned characteristic adaptations, so the magnitude-interval metric for dispositional traits was inapplicable. All entries outside the revisions retained their previous content. Appendix~\ref{app:g-3} lists the revisions by season.

The six revisions concerned three characteristic-adaptation entries: factual and logical priority was revised three times, family-interaction coping twice, and autonomy needs once. Dispositional-trait values and narrative-identity entries remained unchanged. Revisions primarily refined the conditions attached to existing entries, extending applicability to additional types of occasion or specifying ways of responding in particular situations. The revised entries specified when direct confrontation should give way to self-regulation or an indirect response.

\subsubsection{Behavioral Effects of Persona Revisions}
\label{sec:results-revision-effects}

After each revision was committed, independently derived target situations were used to assess whether the revised version expressed the proposed change. The previous and revised versions generated responses in parallel to the same situations. Behavioral realization was rated against four requirements: goals, strategies, concrete actions, and communicative stance. Revisions in Seasons~3, 6, and 7 produced clear behavioral changes, with seasonal mean realization scores increasing from 0.43, 0.55, and 0.34 to 0.96, 1.00, and 0.65, respectively. The pooled gain across six seasons was 0.23, including 16 pairs from Seasons~2 and 4 whose pre-revision scores were already at ceiling. Excluding these pairs yielded a gain of 0.34 across the remaining 32 pairs. Table~\ref{tab:revision-effects} reports the seasonal results.

\begin{table}[htbp]
\centering
\caption{Target-situation behavioral realization before and after persona revision.}
\label{tab:revision-effects}
\small
\renewcommand{\arraystretch}{1.15}
\begin{tabular}{@{}lccc@{}}
\toprule
\textbf{Season} & \textbf{Pre-revision} & \textbf{Post-revision} & \textbf{Gain} \\
\midrule
2 & 1.00 & 1.00 & 0.00 \\
3 & 0.43 & 0.96 & $+0.53$ \\
4 & 1.00 & 1.00 & 0.00 \\
5 & 0.82 & 0.87 & $+0.05$ \\
6 & 0.55 & 1.00 & $+0.45$ \\
7 & 0.34 & 0.65 & $+0.32$ \\
\bottomrule
\end{tabular}
\end{table}

\noindent\textit{Note.} Pre- and post-revision values are seasonal means of target-situation behavioral realization scores.

Two auxiliary comparisons further examined the sources of the revision effects. The full reflection-and-memory accumulation procedure exceeded the no-episode-level-accumulation condition by 0.36 on target-situation tests across 35 eligible pairs. These were situations in which the ablated version had scored below ceiling and improvement remained possible. The other 13 pairs---five in Season~2 and eight in Season~4---were excluded under the predefined rule because the ablated version was already at ceiling. This difference supports a substantive contribution from episode-level reflection and structured memory accumulation to revision quality. In directional paired judgments, the revised version had a tie-adjusted preference proportion of 0.56 against the revised version without current-season policies (57 wins, 47 ties, and 40 losses). The policy-removed revised version had a corresponding preference proportion of 0.68 against the previous version (76 wins, 44 ties, and 24 losses). These judgments showed a larger directional contribution from belief-entry updates than from the added runtime policies. Taken together, the target-situation gains and auxiliary comparisons indicate a consistent, measurable direction of behavioral change, with a limited overall magnitude.

\subsubsection{Case Study of Controllable Evolution}
\label{sec:results-evolution-case}

The Season~3 revision illustrates the full controllable-evolution procedure. Two episodes ultimately provided support for a revision to the family-interaction coping entry: when family authority intervened or family support was interrupted, the protagonist shifted from direct confrontation to self-regulation and temporary acceptance while continuing to defend his position through factual statements. Review identified a remaining competing explanation for each episode. One response could reflect adherence to an established routine; the other could reflect apprehension about his mother's authority. Given the restricted range of situations and limited cross-situational independence, season-end review classified the evidence as limited and narrowed the revision to situations involving family-authority intervention or interruption of family support. Within this narrower scope, the candidate met the predefined minimum evidence standard and was committed as a conditional proposition appended to the family-interaction coping entry.

The corresponding target-situation analysis showed an observable behavioral change. The situation, derived from the Season~3 revision, described the protagonist's mother being unable to drive him to audit university classes as planned because of emotional difficulties and explicitly asking him to suspend his current academic activities. Before revision, the character continued to assert his needs, contesting the restriction with facts and logic. After revision, he discontinued direct confrontation, regulated his response, and temporarily accepted the situation, while maintaining his position through factual statements or questions about alternative arrangements. Mean behavioral realization across the season's target situations increased from 0.43 to 0.96, the largest seasonal gain.

\section{Discussion}
\label{sec:discussion}

Emergi-PersonaOS organizes persona construction, situational expression, and long-term updating around a shared persona object maintained throughout system operation. Enduring content, current states, and event evidence are linked within the same representation. This design makes three contributions.

First, the three-layer persona representation provides a common representational basis for the full persona lifecycle, with separate representations of enduring beliefs and current states. Task-oriented character modeling organizes information around task requirements \citep{wang2024a}. Emergi-PersonaOS assigns content to dispositional traits, characteristic adaptations, and narrative identity according to psychological categories. These assignments remain consistent across tasks, allowing the same structure to support initial construction, adaptive expression, and long-term updating. The three processes consequently operate on a common object and preserve its representational continuity as control passes between them.

Second, the system translates the situational influence of personality into an executable adaptation mechanism. Situational evidence informs a current state under the active persona beliefs, and that state guides actions and replies. Prior work identifies a trade-off between behavioral flexibility and persona consistency \citep{cai2026}. Emergi-PersonaOS addresses these requirements through separate processes: current states accommodate situational variation within a turn, while changes to enduring beliefs undergo a distinct review of revision eligibility and supporting evidence. This separation also informs the temporal basis of evaluation. Static labels tied to a fixed point in a narrative can introduce evaluation noise as a character develops \citep{wang2024b}. State evaluation in the present system therefore uses pre-response material, linking assessment of the current interaction to the persona state available when the response was generated.

Third, research on life events and personality development is operationalized as an evidence-based mechanism for controllable evolution. Long-term changes enter persona beliefs through change attribution, meaning-making, and behavioral testing. Reviewability is implemented through evidence gating, traceability through version relationships and provenance records, and rejectability through candidate handling that permits rejection while preserving eligibility for reassessment. Long-term change is thus represented as a sequence of inspectable operations \citep{tang2026}. Every proposed modification enters through a candidate record, and source checks before activation make the basis of each belief available for inspection \citep{choi2026}. Season-end evidence sets and behavioral tests relate revision outcomes to accumulated experience and its developmental implications \citep{wang2026b}. Separate evaluation dimensions and evidence-sensitive fidelity criteria provide an interpretable basis for comparing versions \citep{tu2024}. Reflection and memory supply candidate material, consistent with their roles in experience-based agent architectures \citep{park2023}. Proposed changes derived from these outputs retain candidate status until the evolution procedure establishes eligibility, allowing rejection and reassessment throughout revision.

Four limitations define the scope of the current implementation. First, all formal ratings were produced by evaluators with fixed generation settings and therefore depend on their judgment tendencies. Randomized labels and blinded comparisons address condition-identification effects, but model dependence remains, particularly because the reliability and validity of LLM personality measurements vary with administration conditions \citep{tosato2025}. Human evaluation has yet to enter the formal procedure, and its implications for absolute scores and condition rankings have not been quantified. Future work will measure evaluator-model dependence and use human-rated reference samples to calibrate absolute scores.

Second, the material follows a single fictional character within one longitudinal narrative. The findings concern system operation for this character and sequence of events. Extension to personalization involving real users requires separate investigation using real-person data, appropriate authorization, and ethics review. Future work will apply the revision procedure to multiple characters to examine individual differences and differentiated outcomes, and will broaden the evaluation of persona change to include calibration of change magnitude.

Third, every stage used the same underlying model: persona construction, longitudinal execution, season-end review, reference generation, and evaluation. The effects of separating the generation and evaluation models or replacing the underlying model remain untested. Future work will introduce this model-level separation and examine sensitivity to the model used.

Fourth, persona revisions were reviewed only at season-end boundaries. Changes developing within a season were retained as monitoring records and matters for further observation. Event-level revision and its evidence rules have yet to be implemented. Future work will introduce event-level review boundaries with corresponding evidence requirements and tests, and will distinguish supporting evidence, counterevidence, scope-boundary evidence, and source credibility more explicitly.

\section{Conclusion}
\label{sec:conclusion}

Motivated by the prospect of symbiosis between humans and digital beings, we present Emergi-PersonaOS, a persona agent operating system that manages a persistent persona object throughout agent operation. A unified representation and continuous execution process support persona construction, situational expression, and long-term updating. The system organizes dispositional traits, characteristic adaptations, and narrative identity into a three-layer persona representation, separating relatively enduring persona beliefs from the current persona state. During situational adaptation, the active belief version remains fixed while situational evidence collection and three-layer state inference produce a state that guides actions and replies. Differences in expression are represented at an inspectable state level. Over longer periods, event records inform revision candidates through change attribution, meaning-making, and behavioral testing. Candidates take effect as new belief versions after meeting predefined evidence standards within an explicit scope; rejected or deferred candidates remain eligible for reassessment. Both processes operate on the same persona object, connecting enduring content, current expression, and development through a shared representation and version history. An end-to-end demonstration using character-centered material from all 141 episodes of \textit{Young Sheldon} showed that the representation and the two processes could operate together as designed over an extended run, preserving the links among enduring persona content, current states, and event evidence.

\phantomsection
\addcontentsline{toc}{section}{References}
\bibliographystyle{plainnat}
\bibliography{references}

\clearpage
\appendix
\section{Data Use and Research Ethics}
\label{app:a}

This study uses textual material from a publicly released television series to investigate a fictional character's persona representation, situational expression, and long-term updating. The research concerns fictional dialogue tasks and excludes psychometric assessment of real participants, private communications, and sensitive identity information. System outputs are computational interpretations of the fictional-character material, and the findings apply to the character-dialogue tasks specified in this paper.

The main text and appendices include only the short dialogue excerpts needed for the analysis, identified by episode and evidence location. Replication materials exclude continuous scripts, complete episode dialogue, and text collections that could substitute for the original work. Materials provided for replication comprise episode identifiers, event indices, evaluation rules, prompts, run manifests, and necessary excerpts that do not allow reconstruction of the original work.

Persona inference uses only the textual evidence permitted by the study design. Plot synopses locate events, and scene descriptions establish relationships and temporal stages. Physical performance, vocal delivery, acting style, and camera information are excluded from persona judgments. Narration, evaluations by other characters, and ambiguously attributed speech are recorded separately during material preparation. Material containing future information or unclear provenance is excluded from pre-response state inference and season-end revision evidence.

Model-generated persona states, character responses, reflections, and revisions are stored separately from the source material. Derived interpretations retain their status as inferences, and every long-term persona modification must be traceable to the original episode, event, and dialogue identifiers.

The numerical and textual values in the three-layer beliefs are internal system representations of a fictional character. They are not interpreted as psychometric trait scores or as assessments of real individuals. All longitudinal evidence comes from one fictional character in one series, so the persona-evolution findings concern system operation for that character over the specified material sequence. Generalization to human personality development requires separate investigation.

\section{Additional Implementation Details}
\label{app:b}

\subsection{Implementation of Comparison Conditions and Behavioral-Test Versions}
\label{app:b-1}

The five situational-adaptation conditions and the no-episode-level-accumulation condition are implemented through input assembly. Section~\ref{sec:experiments} defines their available information and shared settings; this appendix specifies the assembly procedure. For long-term updating, the no-episode-level-accumulation condition assembles current-season source material and state-monitoring records for season-end review, excluding structured memory entries and episode-level reflection conclusions.

The three behavioral-test versions differ in their belief versions and policy sets. The previous version uses the persona version from the preceding season-end checkpoint with all attached policies. The revised version uses the newly committed beliefs, with policies compiled for the current season plus inherited policies. The revised version without current-season policies uses the same new beliefs but retains only inherited policies. This comparison varies the added behavioral policies while holding the revised belief entries fixed. In the behavioral-generation input, every characteristic-adaptation and narrative-identity entry includes an explicit runtime-policy slot. An empty value is supplied when no policy is available, preventing entry identifiers from being mistaken for policy identifiers.

\subsection{Compilation and Attachment of Runtime Behavioral Policies}
\label{app:b-2}

After a season-end candidate passes evidence and field review, the system constructs a candidate belief version. For revisions requiring behavioral policies, the new propositions are compiled into corresponding runtime behavioral policies. The candidate beliefs and any applicable policies together constitute the version evaluated in frozen copies. A policy specifies trigger conditions, policy-selection conditions, policy rules (name, action template, communicative stance, and priority), scope of applicability, exclusion conditions, and fallback behavior. Trigger cues must be observable in the current scene and expressed at the topic-category granularity of the candidate's scope. A trigger that presupposes interaction history cannot be satisfied in a first encounter; programmatic checks therefore reject such triggers during compilation and request revision. The policy's scope is copied verbatim from the candidate revision and limited to the smallest set supported by the new evidence. At least one out-of-scope situation must be specified as an exclusion condition.

When a candidate requires runtime policies, the compiled policies are first attached to the corresponding candidate belief entries in frozen copies. After behavioral testing succeeds, the belief version, reconciled state, and applicable policies are committed together at a turn boundary, with required policies written to the formal run directory. Candidates and policies that fail testing remain outside that directory.

Programmatic checks depend on whether the candidate requires runtime policies. For candidates that do, the system checks that a corresponding policy has been generated before the commit stage; that compilation is complete and consistent with the candidate's scope, triggers, and exclusions; that the policy is attached to the appropriate belief entry in the frozen copy, so the test evaluates the complete candidate; and that the policy record is written to the run directory together with the new belief version at commit time. Failure of any applicable check blocks activation and is recorded with its reason. For candidates requiring no compiled policy, policy-related checks are marked inapplicable, and testing and commitment proceed using the candidate beliefs and reconciled state. Existing policies must be preserved when the policy directory is reconstructed to resume execution across seasons.

\subsection{Implementation Limitations}
\label{app:b-3}

Three implementation limitations warrant attention. First, prompts define module behavior, so changes in wording may affect outputs; prompt versions were fixed before the formal run to limit this variation. Second, modules operate sequentially, allowing upstream state errors to propagate into response generation and evaluation. Episode-level monitoring and programmatic checks intercept only some of these errors. Third, runtime policies express the behavioral implications of revisions as conditional rules, which cover only part of the ways in which persona content can influence behavior. In situations where a policy is not triggered, the revision acts through the belief entry itself.

\section{Persona Representation and Construct Taxonomy}
\label{app:c}

\subsection{Examples of Core Data Objects}
\label{app:c-1}

Persona beliefs are stored as versions. Dispositional traits are represented by numerical Big Five entries, while characteristic adaptations and narrative identity use textual entries. Entries include evidence identifiers, scopes of applicability, and runtime-policy slots. Two examples below illustrate these data structures. The first shows the fields of a characteristic-adaptation belief entry.

\begin{verbatim}
{
  "entry_id": "L2_01",
  "construct_id": "need_xxx",
  "text": "Entry text generated from Season 1 material using source evidence",
  "applicability_scope": [
    "Scope of applicability of the entry"
  ],
  "evidence_ids": [
    "S01E03_Q12"
  ],
  "epistemic_status": "supported",
  "runtime_policies": []
}
\end{verbatim}

The second example shows a runtime behavioral policy compiled at the end of Season~5, with one trigger condition excerpted.

\begin{verbatim}
{
  "policy_id": "RP_S05_L2_06",
  "trigger_conditions": [
    {
      "trigger_id": "T01",
      "cues": [
        "The scene involves a collective church ritual or a similar religious ritual",
        "Sheldon is requested or scheduled to participate physically in the ritual",
        "The physical ritual lacks a verifiable factual or logical basis"
      ],
      "threat_type": "Requirement for physical participation in a religious ritual",
      "interpersonal_context": [
        "Family members",
        "Members of a religious group",
        "Church congregation"
      ],
      "required_scope_match": [
        "Collective church rituals",
        "Requirements for physical participation in religious rituals"
      ]
    }
  ]
}
\end{verbatim}

\subsection{Predefined Construct Taxonomy}
\label{app:c-2}

The construct taxonomy has two levels: general categories and character-specific instances. Table~\ref{tab:construct-taxonomy} defines the general categories assigned to each persona layer and applies across characters. Before formal execution, the character-specific taxonomy was constructed and fixed using these categories and the character's Season~1 material. L1 uses the five Big Five domains. The system generates specific characteristic-adaptation and narrative-identity entries from Season~1 and assigns them to the corresponding construct categories. Entry text, numerical values, evidence, and applicability scopes are generated by the system. Constructs lacking sufficient evidence are assigned an unknown status, with the limitation explained in the entry text. New construct categories outside the taxonomy are excluded.

\begin{table}[htbp]
\centering
\caption{Construct taxonomy.}
\label{tab:construct-taxonomy}
\small
\renewcommand{\arraystretch}{1.15}
\setlength{\personatablewidth}{\dimexpr\linewidth-4\tabcolsep\relax}
\begin{tabular}{@{}>{\raggedright\arraybackslash}p{0.28\personatablewidth}>{\raggedright\arraybackslash}p{0.24\personatablewidth}>{\raggedright\arraybackslash}p{0.48\personatablewidth}@{}}
\toprule
\textbf{Layer} & \textbf{Construct category} & \textbf{Meaning} \\
\midrule
L3: Narrative identity & Overall identity & Core basis of self-identification \\[4pt]
L3: Narrative identity & Imago prototypes & Prototypical images used in self-understanding \\[4pt]
L3: Narrative identity & Life course & Central narrative themes of the individual's life course \\[4pt]
L2: Characteristic adaptations & Needs & Enduring motivational content that the individual seeks or avoids \\[4pt]
L2: Characteristic adaptations & Values & Core standards for judging objects and behavior \\[4pt]
L2: Characteristic adaptations & Goals & Action outcomes pursued over a specified time horizon \\[4pt]
L2: Characteristic adaptations & Strategies & Ways of acting to pursue goals or respond to situations \\[4pt]
L1: Dispositional traits & Personality traits & General behavioral tendencies across situations, represented by the Big Five domains \\
\bottomrule
\end{tabular}
\end{table}

\section{Material Construction, Data Splits, and Evaluation Nodes}
\label{app:d}

\subsection{Material Construction}
\label{app:d-1}

For each episode, the plot synopsis and dialogue in original order are used to construct a character-centered narrative, three to eight key events, key dialogue, information gaps, and provenance records. Key dialogue is preserved verbatim, with original line numbers and its relationship to the target character recorded. Ambiguous speaker attribution, missing event outcomes, and unclear correspondence between synopsis and dialogue are registered as information gaps. Each event is divided into pre-response and post-response material. Pre-response evidence is limited to the current event's situational evidence and previously available beliefs and memories; post-response material is used for monitoring, reflection, and evaluation.

Material quality checks cover episode identifiers and broadcast order, the existence of evidence identifiers, matching against the original dialogue, and the presence of information outside the supplied input. Only records that pass both structural checks and LLM content review enter the formal material set. Source files, prompt and schema versions, and material fingerprints are retained.

\subsection{Data Splits}
\label{app:d-2}

Initialization and longitudinal-execution sets are divided by season. Situational-adaptation nodes and cross-episode memory-check situations are fixed before condition outputs are generated. Post-commit behavioral-test situations are derived by a separate designer after the system commits the season's revision, and the derivation process is withheld from the system.

\begin{table}[htbp]
\centering
\caption{Material sets and data splits.}
\label{tab:data-splits}
\small
\renewcommand{\arraystretch}{1.15}
\setlength{\personatablewidth}{\dimexpr\linewidth-4\tabcolsep\relax}
\begin{tabular}{@{}>{\raggedright\arraybackslash}p{0.32\personatablewidth}>{\raggedright\arraybackslash}p{0.28\personatablewidth}>{\raggedright\arraybackslash}p{0.40\personatablewidth}@{}}
\toprule
\textbf{Material set} & \textbf{Scope} & \textbf{Primary use} \\
\midrule
Initialization set & Season~1 & Construct initial three-layer persona beliefs \\[4pt]
Longitudinal-execution set & Seasons~2--7 & Sequential execution and season-end review \\[4pt]
Situational-adaptation nodes & Seasons~2--7 & Condition comparisons and detailed ratings \\[4pt]
Cross-episode memory assessment & Associated with evaluation nodes & Assess cross-episode and cross-season memory \\[4pt]
Behavioral-test situations & Derived after revision & Assess target realization and scope preservation \\
\bottomrule
\end{tabular}
\end{table}

\subsection{Selection of Situational-Adaptation Evaluation Nodes}
\label{app:d-3}

Candidate nodes are screened for continuous speech by the target character, an identifiable pre-response boundary, locatable key dialogue, relationships that can be established from the available material, and an observable post-response outcome. Eligible candidates are listed in episode order. The final selection provides coverage across relationship categories and seasons.

Each node specifies its episode and event identifiers, pre-response dialogue span, source-response span, relationship category, associated cross-episode memory-check situations, available historical evidence, and excluded information. Evaluation uses the registered spans without expanding the context during rating. Target constructs are drawn from the character-specific taxonomy compiled before formal execution under the general framework in Appendix~\ref{app:c-2}. The mapping between entry identifiers and constructs is fixed before condition outputs are generated.

\subsection{Season-End Evidence Sets and Cross-Episode Memory Assessment}
\label{app:d-4}

Each season-end evidence set contains all valid events, dialogue evidence, state trajectories, monitoring records, and reflection records from that season.

The separate cross-episode memory assessment contains 12 items covering cross-episode event recall, cross-season event recall, same-event interlocutor discrimination, similar-experience discrimination, and false-event rejection. The two recall categories test whether prior events can be retrieved correctly in later situations within or across seasons. Interlocutor discrimination tests whether a memory is bound to the correct interaction partner; similar-experience discrimination tests whether a similar but irrelevant experience is mistakenly cited; and false-event rejection tests whether the system rejects an event that never occurred. Items are administered to the full system and the no-cross-episode-memory condition, with their score difference used to estimate the contribution of memory.

At natural situational nodes, memory-use checks compare the memories actually used in each condition output against the memories identified as necessary by the reference procedure, reporting hit and misuse rates. Relevance to the current judgment is classified as strong, weak, or absent. Nodes requiring historical memory are evaluated for hits and misuse; routine retrieval at nodes requiring no historical memory is recorded separately. The two groups are summarized separately.

\section{Evaluation Metrics, Rating Criteria, and Computational Definitions}
\label{app:e}

\subsection{Structural Compliance Ratings}
\label{app:e-1}

Structural compliance is rated at the episode level. Outputs for situational organization, persona execution, state monitoring, memory formation, reflection, and candidate accumulation are assessed semantically on eight dimensions, each scored from 0 to 2. A score of 0 denotes a clear violation, 1 partial compliance or limited material, and 2 full compliance.

\begin{table}[htbp]
\centering
\caption{Structural compliance dimensions.}
\label{tab:compliance-dimensions}
\small
\renewcommand{\arraystretch}{1.15}
\setlength{\personatablewidth}{\dimexpr\linewidth-4\tabcolsep\relax}
\begin{tabular}{@{}>{\raggedright\arraybackslash}p{0.28\personatablewidth}>{\raggedright\arraybackslash}p{0.72\personatablewidth}@{}}
\toprule
\textbf{Dimension} & \textbf{Check} \\
\midrule
Temporal separation & Does pre-response inference use post-response material from the same episode? \\[4pt]
Event grounding & Are state claims supported within the current event? \\[4pt]
Layer assignment & Is content assigned to the appropriate dispositional-trait, characteristic-adaptation, or narrative-identity layer? \\[4pt]
State sparsity & Is the state limited to content activated in the current situation? \\[4pt]
Evidence traceability & Do cited evidence records exist and refer to the current event? \\[4pt]
Source separation & Are system predictions, source observations, and derived interpretations distinguished? \\[4pt]
Reflection appropriateness & Are reflection conclusions proportionate to the strength of the evidence? \\[4pt]
Candidate safety & Does the candidate specify a bounded scope and address competing explanations? \\
\bottomrule
\end{tabular}
\end{table}

Six critical-error types are recorded: future information entering pre-response inference, source conflation in which a derived interpretation is recorded as a source fact, unauthorized direct changes to enduring persona content during episode processing, invalid evidence identifiers, unsupported scope expansion, and incorrect base-version relationships. Critical errors are assessed separately from the numerical ratings. Any critical error causes the unit to fail, and full structural compliance requires all units to pass. Programmatic-check outputs accompany the materials to help evaluators locate potential problems; the source events remain the factual reference. Section~\ref{sec:structural-compliance} reports critical errors and the four groups of programmatic checks for the formal run.

\subsection{Character-Response Ratings}
\label{app:e-2}

Character responses are rated on four dimensions, each from 0 to 5. Personality fidelity measures consistency with the persona version and current state: high scores require clear expression of the relevant persona content, while low scores indicate substantial conflict with core content. Situational coherence measures the handling of characters, relationships, events, and topics, with high scores for complete and accurate treatment and low scores for clear conflicts with the situation.

State--behavior consistency measures how state content is expressed in behavior. High scores require identity, goals, strategies, and traits to jointly inform a coherent course of action; low scores indicate behavior opposed to the current state. Dialogue continuity measures connection to the preceding dialogue and adherence to the available information. High scores require coherence and observance of temporal boundaries; low scores indicate an inability to continue the dialogue or the use of critical future information. Raw ratings, reasons, and evidence identifiers are recorded for each evaluation. Inter-rater agreement and within-condition internal consistency are reported with the results. Paired comparisons use the same node and metric.

\subsection{Long-Term Persona Update Ratings}
\label{app:e-3}

System and reference decisions are recorded in a three-category table: revise, retain, or defer. Evidence-strength categories are defined in Section~\ref{sec:evaluation-tasks}. A candidate supported by a single episode or partly consistent evidence must first be limited to a narrower scope. It can be committed only if it then meets the predefined minimum evidence standard; otherwise, it is deferred. Supporting evidence for a revision must remain traceable to source events \citep{choi2026}. Content evaluation covers the target layer, revision direction, and preservation of unmodified content. Magnitude-interval overlap is additionally evaluated when the revision concerns dispositional traits.

For characteristic-adaptation and narrative-identity entries, semantic change and scope of applicability are reported separately. Unmodified content is checked for numerical drift in L1 and entry-level text changes in L2 and L3. Minor changes alter wording while preserving persona meaning; substantive changes add, delete, or rewrite content in ways that change its meaning, direction, or applicable target. Update evaluation also examines coverage, preservation of existing content, and fidelity to source material \citep{yang2026}.

\begin{table}[htbp]
\centering
\caption{Recording revision-content evaluations.}
\label{tab:revision-content}
\small
\renewcommand{\arraystretch}{1.15}
\setlength{\personatablewidth}{\dimexpr\linewidth-4\tabcolsep\relax}
\begin{tabular}{@{}>{\raggedright\arraybackslash}p{0.36\personatablewidth}>{\raggedright\arraybackslash}p{0.64\personatablewidth}@{}}
\toprule
\textbf{Evaluation item} & \textbf{Recorded categories or measure} \\
\midrule
Target layer & Correct; partly correct; incorrect \\[4pt]
Revision direction & Consistent; partly consistent; opposite; unspecified by the reference \\[4pt]
Magnitude-interval overlap & Degree of overlap between the revision interval and the expected reference interval \\[4pt]
Preservation of unmodified content & Preserved; minor change; substantive change \\
\bottomrule
\end{tabular}
\end{table}

\subsection{Behavioral Tests and Directional Paired Judgments}
\label{app:e-4}

Behavioral testing uses target and out-of-scope situations. Target tests assess expression of the proposed revision through surface paraphrases of the same situation and transfer to semantically related situations, with at least four of each type per season and at least eight in total. Out-of-scope tests cover three reasons why a revision should remain inactive: scope mismatch, insufficient trigger cues, and satisfied exclusion conditions. Each reason is represented by at least one situation. These tests check inapplicability and do not produce a separate quantitative metric. Target-situation behavioral realization is rated on goals, strategies, concrete actions, and communicative stance, each from 0 to 5. The four ratings are averaged with equal weights and divided by 5, yielding a realization score from 0 to 1. The seasonal score is the arithmetic mean across that season's target situations. The formal run used eight target situations and four to six out-of-scope situations per season.

A separate designer derives the post-commit test situations from the revision actually submitted for the current season. The designer can access the submission and the entries before and after modification, but has no access to generated behavioral outputs or the compiled policy text. Programmatic checks verify that each situation matches the submission's layer, entry, direction, and scope. Situations that depart from the submission are rejected and regenerated.

The test dimensions draw on behavioral approaches to personality change, in which situational behavior, delayed reassessment, and persistence of direction are used to examine the expression and continuity of change \citep{wang2026b}. Separate checks of target realization, paraphrase generalization, and specificity to related content are informed by model-editing evaluation \citep{meng2022}. Both intended behavioral change and preservation of behavior on other inputs are relevant evaluation criteria \citep{yao2023}.

Directional paired judgments assess how revisions are expressed in behavior. Given the direction of a season's committed revision, the evaluator compares two anonymized outputs for the same test situation and selects the one that better expresses that direction. The revised version is compared with the revised version without current-season policies to examine the policies' contribution. The policy-removed revised version is then compared with the previous version to examine the contribution of belief entries. Each season has eight test situations, and each sampling repetition includes both comparisons. Left--right order is randomized for each pair, condition identities are hidden, and ties receive a weight of 0.5. Results are summarized within and across seasons. The anonymized policy comparison is repeated three times and includes left--right swaps.

\subsection{Statistical Analysis}
\label{app:e-5}

Situational-adaptation comparisons use the evaluation node as the pairing unit. Condition means and node-level paired differences are reported for each metric, with point estimates for proportional measures such as paired preferences. Inter-rater agreement is calculated as ICC(2,1), using a two-way random-effects model and treating each sample-by-evaluator rating as a single measurement. Internal consistency of the overall character-response score is reported separately for each condition using Cronbach's~$\alpha$, with the four behavioral dimensions as items. Paired preference proportions assign ties a weight of 0.5 and are summarized within and across seasons. The three repetitions of the anonymized policy comparison include left--right swaps to address presentation-order effects. Statistics are reported separately under the predefined analysis rules. No unplanned weighted composite metrics or post hoc subgroup analyses are introduced.

\section{Generation and Evaluation Settings}
\label{app:f}

\subsection{Model and Parameters}
\label{app:f-1}

The formal run accessed \texttt{deepseek-v4-pro-0813} through a unified OpenAI-compatible interface. All tasks used temperature 0.2, with no random seed specified. The maximum output length was generally 6,000 tokens and varied by task type. Retries retained the same experimental input, prompt, schema, and generation settings, and every attempt was recorded separately. Model identifiers, prompt versions, and interface configurations were fixed before formal execution.

\subsection{LLM Reference Generation and Evaluation}
\label{app:f-2}

The situational-node reference generator reads pre-response material, available historical evidence, and the three-layer persona definitions, and returns a reference state, an acceptable range, and evidence identifiers. The character-response reference generator reads pre-response material, the character's actual source response, and subsequent dialogue to establish the basis for the four response ratings. The season-end reference generator uses the previous persona version and current-season source evidence to produce a reference decision and its evidential scope. When the material supports multiple interpretations, acceptable directions and their corresponding evidence are recorded together.

The evaluator receives the fixed rating criteria, corresponding references, permitted evidence, and condition outputs with randomized labels. It returns dimension-level scores, reasons, evidence identifiers, and inapplicability markers. Each evaluation unit is scored once per evaluator call under fixed generation settings; repeated measurements are supplied by separate response samples and evaluator instances.

\subsection{Condition Randomization and Blinded Evaluation}
\label{app:f-3}

A local program assigns randomized labels to the condition outputs within each evaluation unit. Label mappings remain hidden from evaluators until rating is complete. Each node produces three response samples, each rated by three separate evaluator instances with independently randomized labels. Label assignment is crossed across samples and evaluators. Once condition names are restored, the program checks label completeness and the number of conditions for each unit. Units with missing or duplicated labels are excluded from effect calculations and recorded as evaluation failures.

\subsection{Evaluator Admission and Precalibration}
\label{app:f-4}

Evaluators must pass an admission check before formal rating. The check uses ten control pairs containing deliberately inserted errors, including fabricated evidence identifiers, temporal violations, and condition-identity leakage. An evaluator enters formal evaluation only after reaching the predefined correct-identification threshold. Admission checks are conducted in three batches: after longitudinal execution, before the strict evaluation stage, and before paired comparisons.

\section{Extended Results and Case-Level Examples}
\label{app:g}

\subsection{Node-Level Paired Differences in Situational Adaptation}
\label{app:g-1}

A node-level paired difference is the difference between two conditions on the same metric at the same situational-adaptation node. Pairing within nodes accounts for differences in node difficulty when comparing conditions. The 48 evaluation nodes comprise 24 family situations and 24 school or peer situations. Table~\ref{tab:grouped-results} summarizes condition means and paired differences by relationship category.

\subsection{Grouped Summaries}
\label{app:g-2}

Table~\ref{tab:grouped-results} reports condition means and paired differences for family situations and school or peer situations.

\begin{table}[htbp]
\centering
\caption{Grouped results for situational-adaptation baselines and ablations.}
\label{tab:grouped-results}
\small
\renewcommand{\arraystretch}{1.15}
\setlength{\personatablewidth}{\dimexpr\linewidth-4\tabcolsep\relax}
\begin{tabular}{@{}>{\raggedright\arraybackslash}p{0.28\personatablewidth}cccc@{}}
\toprule
\textbf{Condition} & \textbf{Family mean} & \textbf{School/peer mean} & \textbf{Family $\Delta$} & \textbf{School/peer $\Delta$} \\
\midrule
Full system & 4.41 & 4.56 & --- & --- \\
Dialogue-only & 2.79 & 2.76 & 1.62 & 1.80 \\
Static persona & 4.38 & 4.51 & 0.03 & 0.04 \\
L1 only & 3.69 & 3.82 & 0.71 & 0.73 \\
No cross-episode memory & 4.33 & 4.50 & 0.02 & 0.03 \\
\bottomrule
\end{tabular}
\end{table}

\noindent\textit{Note.} $\Delta$ denotes the paired difference, calculated as full system minus the comparison condition at the same node. For the no-cross-episode-memory condition, differences use the predefined restricted set of 21 nodes identified as requiring memory: nine family nodes and 12 school or peer nodes. Other comparisons use all 24 nodes in each relationship category.

\subsection{Season-Level Decisions and Revision Content}
\label{app:g-3}

At all six season-end checkpoints, both the system and the reference procedure decided to commit a revision. Evidence was consistent across multiple episodes in Seasons~2 and 4, and came from a single episode or was only partly consistent in the other four seasons.

Season~2 added a condition to the factual-and-logical-priority entry, specifying that factual correctness should retain priority under social pressure in public or family social situations. Season~3 added a condition to the family-interaction coping entry: intervention by family authority or interruption of family support should elicit self-regulation and temporary acceptance in place of direct confrontation. Season~4 extended the factual-and-logical-priority entry to knowledge-related public and social situations. Season~5 added fact-prioritizing expression during collective church rituals. Season~6 revised the autonomy-needs entry to specify responses to frustrated autonomy or authority. Season~7 revised the family-interaction coping entry to permit indirect or symbolic emotional expression and participation in rituals.

\subsection{Run Failures and Missing Evaluation Items}
\label{app:g-4}

Evaluation units excluded from effect calculations, failure categories, and their handling are reported separately from effect scores. All formal situational-adaptation nodes, cross-episode memory items, and season-end checkpoints with their test situations entered evaluation. No unit was excluded because of a run failure or missing evaluation.

\section{Run Records, Error Categories, and Replication Information}
\label{app:h}

\subsection{Run Records}
\label{app:h-1}

Each formal run records its material scope, condition, model identifier, prompt version, generation settings, evaluation settings, and execution time. Each model call records its request fingerprint, raw response, parsed output, elapsed time, retry count, and token usage. SHA-256 hashes are calculated separately for the materials, prompts, schemas, and experimental settings. Identical model identifiers, task roles, inputs, prompts, schemas, and generation settings produce the same request fingerprint. Results that have passed structural and content checks may be reused. Failed and superseded results retain their original records and reasons for replacement.

\subsection{Model-Call Statistics}
\label{app:h-2}

The formal run recorded 2,327 model calls, of which 2,325 were accepted and two abandoned. There were 165 retries following failures.

\subsection{Error Categories}
\label{app:h-3}

Errors are classified into six categories: material, temporal, persona-representation, evidence, version, and evaluation errors. Material errors include incorrect dialogue attribution, incorrect dialogue order, and missing necessary context. Temporal errors include premature use of post-response information and use of later-season material in the current season-end decision, including violations intercepted at the input stage. Persona-representation errors include conflation of enduring beliefs with current states, incorrect layer assignment, and unsupported expansion of revision scope. Evidence errors include invalid source identifiers, substitution of generated content for source facts, and omitted counterevidence. Version errors include incorrect base-version relationships, premature candidate activation, inconsistent inputs across compared versions, and loss of runtime behavioral policies. Evaluation errors include condition-label leakage, system outputs entering reference-generation inputs, and misaligned pairing units. Units affected by temporal, evidence, version, or evaluation errors are excluded from effect calculations. Material errors are first checked against source records. All categories report error counts, affected units, and handling outcomes.

\subsection{Replication Materials}
\label{app:h-4}

Replication materials comprise source-material manifests, evaluation nodes and assessment items, season-end evidence sets, comparison-condition specifications, complete prompts, rating criteria, programmatic checks, and statistical scripts. Copyrighted continuous scripts and complete episode dialogue are excluded from the public replication package. Researchers can consult legally available sources using the episode identifiers and evidence locations to verify the original text.

\end{document}